%% file: main_arxiv.tex
\documentclass{article}

\usepackage[preprint]{corl_2026} 
\usepackage{amsmath,amssymb}
\usepackage{bm}
\usepackage{booktabs}
\usepackage{graphicx}
\usepackage{wrapfig}
\usepackage[
  expansion=alltext,
  protrusion=alltext-nott, 
  final 
]{microtype}

\title{HUGS: Guiding Unified Dexterous Grasp Synthesis Across Modes and Scales via Learned Human Priors}

\newcommand{\equalcontrib}{\textsuperscript{*}}
\newcommand{\correspondingauthor}{\textsuperscript{\dag}}

\author{
  \textbf{Mingrui Yu}\equalcontrib \quad
  \textbf{Yongpeng Jiang}\equalcontrib \quad
  \textbf{Yongyi Jia} \quad
  \textbf{Kangchen Lv} \\
  \textbf{Xiangjie Yan} \quad
  \textbf{Li Huang} \quad
  \textbf{Yi Ren} \quad
  \textbf{Xiang Li}\correspondingauthor \\
  Tsinghua University \\
  \equalcontrib Equal contribution. \quad
  \correspondingauthor Corresponding author. \\
  \href{https://hugs-dex.github.io/}{\texttt{https://hugs-dex.github.io/}}
}

\begin{document}
\maketitle

\input{sections/teaser}
\input{sections/abstract}
\input{sections/introduction}

\input{sections/related_work}
\input{sections/method}
\input{sections/results}
\input{sections/conclusion}


\clearpage
\bibliography{main}  

\clearpage
\appendix
\input{sections/appendix}

\end{document}

%% file: sections/teaser.tex

\begin{figure}[h]
    \centering
    \vspace{-1em}
    \includegraphics[width=0.9\linewidth]{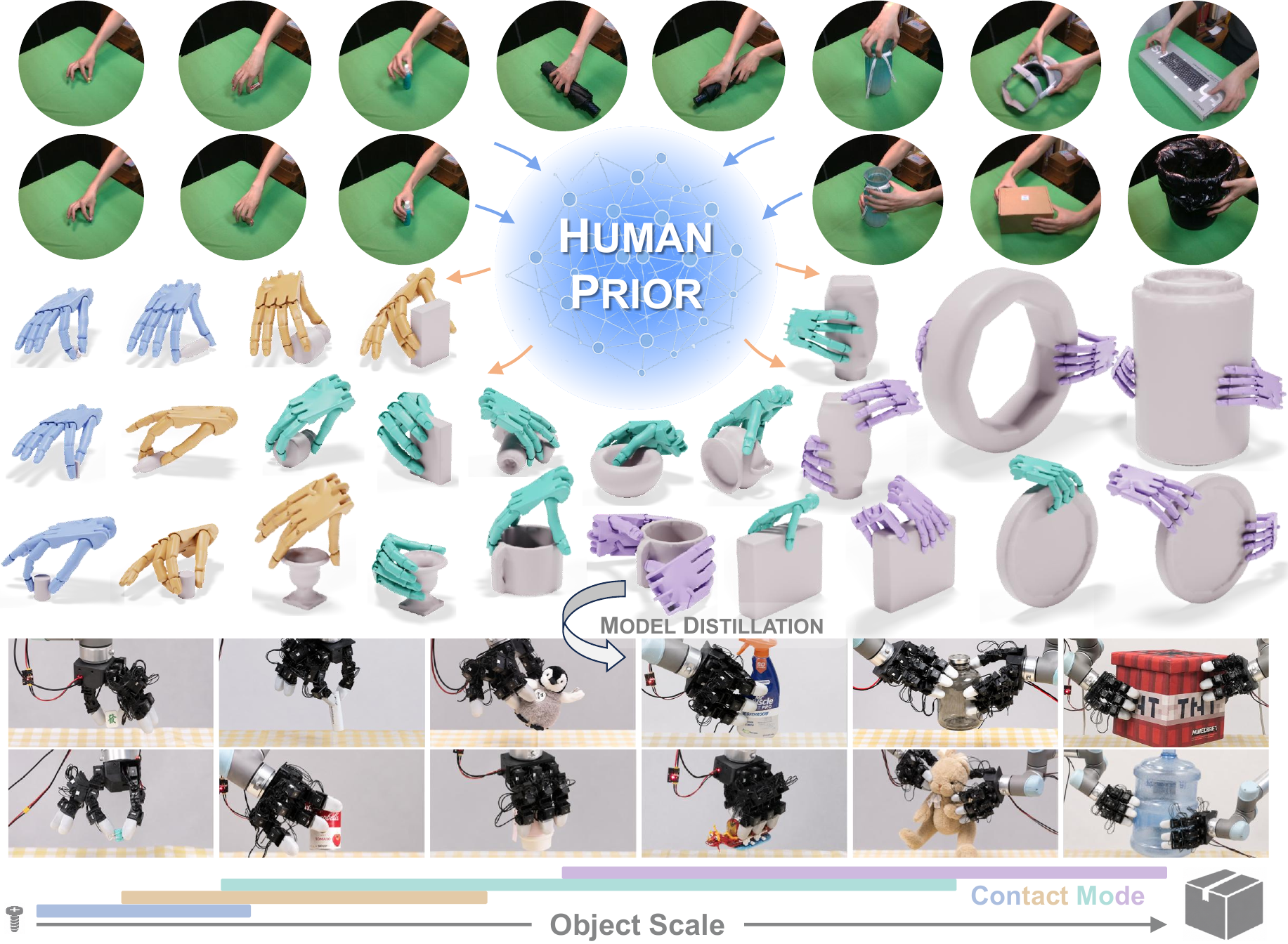}
    \vspace{-0.5em}
    \caption{\textbf{Human priors guide scalable dexterous grasp synthesis across modes and scales.}
    HUGS learns an object-conditioned human prior from a compact self-collected dataset to predict preferred contact modes and wrist initializations.
    Guided by this prior, force-closure-aware optimization synthesizes diverse and stable grasps ranging from two-finger pinches to bimanual grasps.
    }
    \vspace{-0.5em}
    \label{fig:teaser}
\end{figure}


%% file: sections/abstract.tex

\begin{abstract}
Dexterous grasping across diverse object scales requires contact modes ranging from two-finger pinches to bimanual grasps. Existing dexterous grasp synthesis methods reduce the high-dimensional optimization space with manually designed expected contacts and initialization heuristics, which struggle to balance synthesis success rate and diversity.
We present \textbf{HUGS}, a \underline{H}uman-prior-guided framework for \underline{U}nified dexterous \underline{G}rasp \underline{S}ynthesis across modes and scales. Instead of directly retargeting human demonstrations, HUGS learns an object-conditioned human prior that captures human grasp preferences and guides downstream force-closure-aware optimization. The prior is trained on a compact self-collected human grasp dataset with 1.8K grasps over 304 objects, providing broad coverage of object scales and contact modes.
During synthesis, HUGS adaptively proposes contact modes and wrist initializations, substantially improving the balance between contact-mode coverage and synthesis success rate over heuristic-based methods. With HUGS, we synthesize 3.2M robotic grasps over 157K scenes, spanning object half-diagonal lengths from 2~cm to 30~cm and modes from two-finger to bimanual grasps. Models trained on the synthesized dataset autonomously select appropriate contact modes in the real world, enabling grasping from screws to large boxes.
\end{abstract}

\keywords{Dexterous Grasping, Human Priors, Grasp Synthesis}


%% file: sections/introduction.tex

\section{Introduction}
\label{sec:introduction}

Dexterous grasping is inherently multi-mode across object scales, from two-finger precision grasps to coordinated bimanual grasps, and the same object may admit different modes under different conditions. Training a generalizable grasp generation model for such diversity requires massive datasets, yet large-scale real-world collection is impractical, making scalable grasp synthesis essential.

Dexterous grasp synthesis involves a vast search space due to the high degrees of freedom and multi-contact nature.
Existing methods often constrain optimization with manually predefined contact modes, such as single-hand \cite{wang2023dexgraspnet,chen2025bodex} or dual-hand full-finger grasps \cite{shao2024bimanual,lin2026bidexgrasp}, and heuristic wrist pose initializations.
However, these coarse rules fail to exploit object-specific information and struggle to balance synthesis efficiency and grasp diversity: loose heuristics lead to low-quality optimization and inefficient synthesis, while restrictive heuristics severely limit grasp diversity.

Indeed, anthropomorphic robotic hands are designed for human-like manipulation~\cite{billard2019trends}, making human grasps a natural source of knowledge for robots.
Existing methods often retarget human grasps to robotic hands~\cite{wei2024learning,chen2025web2grasp,wang2025scaleadfg}, converting each demonstration into a robot grasp for the same or highly similar object and thereby limiting scalable synthesis over large-scale, diverse object sets.


Our key insight is that we can use human demonstrations to learn generalized and reusable \textbf{object-conditioned human priors}, instead of directly retargeting each human grasp to a robotic hand in a one-to-one manner.
By capturing human preferences for specific object geometry, the prior guides robotic grasp optimization with better global initializations and optimization targets for efficient convergence to higher-quality solutions.
Such a \textbf{Human Prior + Robot Optimization} paradigm offers following advantages: 
\textbf{1) Adaptivity}: compared with hand-crafted heuristics, the learned human prior provides object-aware global guidance that constrains optimization to more suitable regions of search space.
\textbf{2) Scalability}: unlike direct retargeting methods, the learned prior generalizes to unseen objects, enabling scalable synthesis of large and diverse robotic grasp datasets from limited human demonstrations.
and \textbf{3) Physical Plausibility}: force-aware optimization accounts for robot-specific kinematics and task physics, while tolerating slight errors in coarse human prior.

In this work, we present \textbf{HUGS}, a \underline{H}uman-prior-guided framework for \underline{U}nified dexterous \underline{G}rasp \underline{S}ynthesis across modes and scales.
To learn a generalizable human prior from limited data, we abstract human preference into contact modes and wrist poses, the two components most critical for robotic grasp synthesis.
Using a self-collected dataset with 304 objects and 1.8K grasps, we train an object-conditioned generative model that predicts human-preferred grasp configurations.
During synthesis, the learned prior proposes object-adaptive contact modes and wrist initializations, enabling scalable and efficient synthesis of 3.2M grasps over 157K scenes, spanning objects with half-diagonal lengths from 2~cm to 30~cm and contact modes from two-finger to bimanual grasps, forming what is, to our knowledge, the first large-scale dataset with multiple modes per object (Fig. \ref{fig:teaser}). Models trained on the dataset adaptively select suitable contact modes, enabling grasping from screws to large boxes in the real world.
%
Our contributions are highlighted as follows:
\begingroup
\setlength{\leftmargini}{1.0em}
\setlength{\itemsep}{0pt}
\setlength{\parsep}{0pt}
\setlength{\topsep}{0pt}
\begin{enumerate}
    \item \textbf{Human Grasp Dataset Across Modes and Scales}. We collect and publicly release a compact yet diverse human grasp dataset spanning a broad range of object scales and contact modes, providing a strong foundation for learning generalizable human grasp priors.

    \item \textbf{Human-Prior-Guided Unified Grasp Synthesis}. We propose a human-prior-guided framework that learns object-conditioned priors over contact modes and wrist poses for unified dexterous grasp synthesis across scales and modes, eliminating manually designed coarse heuristics.

    \item \textbf{Large-Scale Synthesis and Evaluation}. We demonstrate scalable synthesis of diverse dexterous grasp datasets and systematically evaluate the framework in terms of cross-scale and cross-mode prior prediction, synthesis quality, efficiency, diversity, and real-world grasping potential.
\end{enumerate}
\endgroup

%% file: sections/related_work.tex
\section{Related Work}
\label{sec:related_work}

\textbf{Dexterous Grasp Synthesis.}
Analytical grasp synthesis methods optimize hand configurations with physics-based objectives such as force closure. To reduce the search space of high-DoF multi-contact systems, existing methods typically rely on manually predefined contact regions and heuristic wrist initialization strategies. Most prior work focuses on single-hand full-finger grasps \cite{liu2021synthesizing, wang2023dexgraspnet, chen2025bodex, zurbrugg2025graspqp} or bimanual grasps \cite{shao2024bimanual,lin2026bidexgrasp}.
Different contact modes can be synthesized by changing predefined contact regions, e.g., two-finger \cite{ye2025power, yang2026ultradexgrasp} or three-finger grasps \cite{he2025dexvlg, yang2026ultradexgrasp}. However, these modes are typically manually tied to object-scale ranges, causing each object scale to correspond to only one grasp mode and limiting contact-mode diversity.
In addition, most methods randomly sample wrist poses around the object’s surface \cite{wang2023dexgraspnet, chen2025bodex, shao2024bimanual}. For bimanual grasping, random sampling ignores inter-hand coordination, while strict symmetry constraints restrict diversity \cite{shao2024bimanual, yang2026ultradexgrasp}.
Hand-centric methods \cite{chen2025dexonomy, yin2025lightning} improve diversity and efficiency for single-hand floating grasps, but may be less suitable for bimanual grasping and environment-constrained settings such as tabletop grasping.

\textbf{Human Data for Grasp Synthesis.}
Human demonstrations are widely used for functional grasp synthesis through retargeting-based approaches that transfer human grasps to robotic hands. Some methods directly retarget each human grasp to a robotic grasp \cite{chen2025web2grasp, wei2024grasp}, while others synthesize grasps for similar objects from a small set of human grasp templates \cite{wei2024learning, mao2025universal, huang2025fungrasp, wang2025scaleadfg}.
However, limited by the scale of human functional grasp datasets, these approaches typically support only limited object categories and rely on similarity to existing templates, rather than generalizing to arbitrary objects.
Some works learn object-conditioned contact maps from human data \cite{ma2026contact, wu2025cedex}, but remain limited to single-hand full-finger grasps and require highly accurate human contact predictions that may not align with robotic hand kinematics.
In contrast, our method uses only high-level human priors, namely contact modes and wrist poses, while retaining sufficient flexibility for force-closure grasp optimization compatible with robotic hand kinematics.

\textbf{Online Grasp Generation.}
Synthetic grasp datasets are widely used to train grasp generation networks from partial observations such as images or point clouds for real-time deployment \cite{wei2025dro, weng2024dexdiffuser,xu2023unidexgrasp,he2025dexvlg}, including grasp generation in cluttered environments \cite{zhang2026cadgrasp, zhong2026dexgraspvla, zhang2024dexgraspnet}. 
A few works directly learn grasping policies with reinforcement learning, bypassing explicit grasp pose planning \cite{yuan2025demograsp, zhang2025robustdexgrasp, chen2025clutterdexgrasp, lum2025dextrah,singh2024dextrah}; however, they often suffer from limited grasp diversity and unnatural finger motions.



%% file: sections/method.tex

\section{Method}
\label{sec:method}

\subsection{Problem Formulation}

\textbf{Grasp Optimization Problem.}
Let $o$ denote the target object with surface $\mathcal{S}_o$.
For hand $h \in \{L,R\}$, let $\bm{q}^h \in \mathbb{R}^{d_q}$ be the joint configuration and $\bm{T}^h=(\bm{R}^h,\bm{t}^h)\in SE(3)$ be the wrist pose, with bimanual states $\bm{Q}=(\bm{q}^L,\bm{q}^R)$ and $\bm{T}=(\bm{T}^L,\bm{T}^R)$.
A grasp $g=(\bm{Q},\bm{T},c)$ uses contact configuration $c$ to specify which hand regions contact the object, and grasp synthesis models $\pi(g \mid o)=\pi(\bm{Q},\bm{T},c \mid o)$.
Active contact regions $c$ induce contact points $\{\bm{p}_i\}_{i=1}^{N_c}$ with normals $\{\bm{n}_i\}_{i=1}^{N_c}$.
Under point contact with Coulomb friction, each contact force satisfies $\bm{f}_i \in \mathcal{F}_i=\{\bm{f}_i \mid \|\bm{f}_i^t\| \leq \mu f_i^n,\ f_i^n \geq 0\}$ and induces wrench $\bm{w}_i=[\bm{f}_i;\ (\bm{p}_i-\bm{x}_{\mathrm{com}})\times \bm{f}_i]$, where $\bm{x}_{\mathrm{com}}$ is the object center of mass.
The grasp wrench space is $\mathcal{G}(g)=\{\sum_{i=1}^{N_c}\bm{w}_i \mid \bm{f}_i\in\mathcal{F}_i\}$, and force closure holds when $0\in\mathrm{int}(\mathcal{G}(g))$.
We formulate grasp optimization as minimizing force-closure energy $\Phi(g)$ under feasibility constraints:
\begin{equation}
\label{eq:grasp_optimization}
\min_{g=(\bm{Q},\bm{T},c)} \Phi(g)
\quad \mathrm{s.t.}\quad
\bm{Q} \in [\bm{Q}_{\min}, \bm{Q}_{\max}],\;
\bm{p}_i \in \mathcal{S}_o\ \forall i,\;
\mathrm{CollisionFree}(g,o).
\end{equation}
Here $\bm{p}_i$ are induced by active regions in $c$; the constraints enforce joint limits, object-surface contacts, and collision-free hand-object and hand-hand configurations.

\textbf{Overview of HUGS.}
In grasp pose optimization, the contact configuration $c$ determines the number and locations of contact points, making the problem hybrid discrete-continuous. Existing methods typically reduce the combinatorial search space with predefined contact regions. Meanwhile, the wrist pose $\bm{T}$ is highly global, and poor initializations often trap local optimization in suboptimal minima. Consequently, optimization quality largely depends on initializing $c$ and $\bm{T}$, while optimizing the hand joint configuration $\bm{Q}$ is relatively straightforward given suitable $c$ and $\bm{T}$.
Existing methods mainly use fixed coarse heuristics to initialize contact configurations and wrist poses, wasting optimization budget on implausible grasp modes while missing object-aware strategies.
HUGS instead learns an object-conditioned human grasp prior to organize the search space before robot-specific optimization.

We adopt a compact discrete contact-mode representation with four coarse modes, selected as prevalent human grasping strategies that current dexterous robot hands can execute:
1) \textbf{Single-Two}, a single-hand two-finger grasp for small objects with limited contact area;
2) \textbf{Single-Three}, a single-hand three-finger grasp, the minimum for force closure under frictional point contacts \cite{lynch2017modern};
3) \textbf{Single-Full}, a denser single-hand full-finger grasp requiring larger accessible object surface area;
and 4) \textbf{Both-Full}, a full-finger bimanual grasp for large objects requiring broader surface coverage.
Based on this contact-mode abstraction, we formulate the grasp synthesis as
\begin{equation}
\pi(\bm{Q},\bm{T},c \mid o)
=
\pi(\bm{Q},\bm{T} \mid \bm{T}_0,c,o)\,
\pi(\bm{T}_0 \mid c,o)\,
\pi(c \mid o),
\end{equation}
where the learned prior predicts $\pi(c \mid o)$ and $\pi(\bm{T}_0 \mid c,o)$ to propose plausible contact modes $c$ and wrist initializations $\bm{T}_0$. Conditioned on these high-level proposals, the robot optimizer solves for $\bm{Q}$ and refines $\bm{T}$ under the force-closure and feasibility constraints in Eq.~\ref{eq:grasp_optimization}.

\subsection{Compact Human Grasp Dataset}

\begin{figure}[t]
    \centering
    \includegraphics[width=\linewidth]{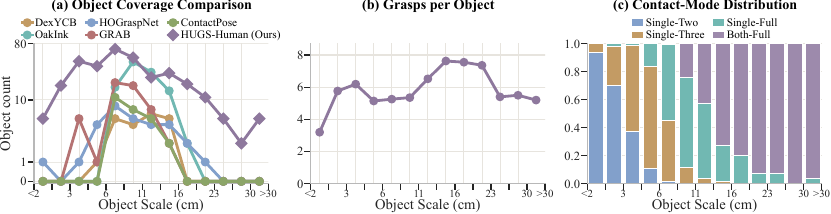}
    \vspace{-1.0em}
    \caption{\textbf{Statistics of HUGS-Human Dataset.} (a) Object-count distribution, compared with existing human grasp datasets (OakInk: only real-world captured objects; log-scaled y-axis and power-scaled x-axis, with absolute-value labels). (b) Average number of HUGS-Human grasps per object. (c) Contact-mode distribution in HUGS-Human.}
    \label{fig:human_dataset_statistics}
    \vspace{-0.5em}
\end{figure}

\textbf{Dataset Motivation.}
Existing human grasp datasets mainly target hand-object reconstruction rather than synthesis guidance.
Most emphasize single-hand grasps with limited scale coverage, such as DexYCB \cite{chao2021dexycb}, OakInk \cite{yang2022oakink}, and HOGraspNet \cite{cho2024dense}; those containing bimanual grasps, such as GRAB \cite{taheri2020grab} and ContactPose \cite{brahmbhatt2020contactpose}, include them only sparsely.
They also do not explicitly consider mode and wrist-pose diversity, making them less suitable for learning priors over $c$ and $\bm{T}_0$ across scales.

\textbf{Self-Collected Dataset.}
We introduce \textbf{HUGS-Human}, a compact human grasp dataset designed to learn the high-level prior used by HUGS.
Participants grasp and lift tabletop objects from multiple supporting poses.
For each posed object, we encourage diverse human-plausible contact modes and wrist poses while avoiding duplicates.
The dataset contains 304 canonical objects, 525 posed objects, and 1.8k distinct human grasps.
The collected grasps reveal three consistent patterns:
1) humans adapt contact modes and wrist poses according to object geometry and scale;
2) a single object can support multiple plausible contact modes, especially when its geometry is asymmetric;
and 3) contact modes and feasible wrist poses are strongly coupled, as certain wrist poses only permit sparse fingertip contacts while others enable stable full-hand grasps on the same object.
These patterns motivate our object-conditioned prior over $c$ and $\bm{T}_0$.
Each collected data sample includes the object mesh, object pose, hand wrist pose, MANO pose \cite{romero2017embodied}, and contact mode. The setup and annotation details are provided in the appendix.
As shown in Fig.~\ref{fig:human_dataset_statistics}, HUGS-Human provides broad object-scale coverage compared with existing datasets, especially for very small and large objects, maintains multiple grasps per object across object scale bins, and captures the scale-dependent transition from sparse single-hand modes to full-hand and bimanual modes.


\begin{figure}[t]
    \centering
    \includegraphics[width=\linewidth]{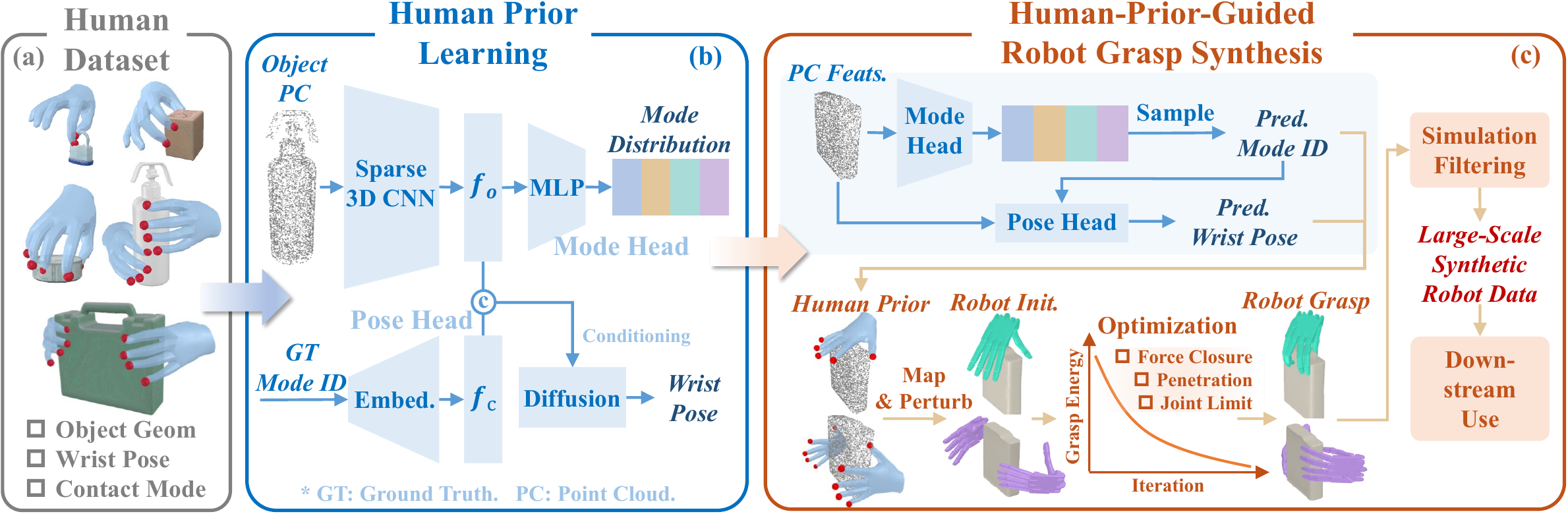}
    \vspace{-1.5em}
    \caption{\textbf{Overview of HUGS.} HUGS learns an object-conditioned human prior from demonstrations with geometry, wrist poses, and contact modes. During synthesis, it predicts contact-mode distributions and wrist initializations for unseen objects, transfers them to robot hands, and optimizes force closure under feasibility constraints to synthesize large-scale robot grasps across modes and scales.}
    \label{fig:method_overview}
    \vspace{-1em}
\end{figure}

\subsection{Object-Conditioned Human Grasp Prior}

\textbf{Prior Formulation.}
We learn object-conditioned distributions over contact modes and wrist poses, denoted by $\pi(c \mid o)$ and $\pi(\bm{T}_0 \mid c,o)$.
Because human and robot wrist frames may differ, $\bm{T}_0$ uses the index metacarpophalangeal (MCP) position for translation and the dorsal-hand wrist orientation for rotation.
We interpret these distributions as a \textbf{synthesis-budget prior}: for each object, they specify how much optimization effort to allocate to each contact mode and wrist-pose region.
Implausible modes receive near-zero budget, while plausible modes share the budget according to human grasp diversity.
This guides synthesis toward diverse plausible grasps with fewer optimization attempts.

\textbf{Architecture and Training.}
As shown in Fig.~\ref{fig:method_overview}(b), the prior takes a centered tabletop-frame object point cloud and encodes it into $\bm{f}_o$ using a MinkowskiEngine Sparse3DConv backbone~\cite{choy20194d}.
A mode head predicts $\pi(c \mid o)$ with an MLP and softmax, supervised by the empirical contact-mode distribution of demonstrations for the same posed object using soft-label cross entropy.
A mode-conditioned diffusion head models $\pi(\bm{T}_0 \mid c,o)$ by conditioning on $\bm{f}_v=[\bm{f}_o,\bm{f}_c]$, where $\bm{f}_c$ is a learnable mode embedding, and is trained on individual data samples with contact-mode labels and human wrist poses.
We use random vertical rotations and point-cloud noise for augmentation.

\subsection{Human-Prior-Guided Grasp Synthesis}

\textbf{Synthesis Overview.}
During grasp synthesis, the learned human prior predicts $\pi(c \mid o)$ and $\pi(\bm{T}_0 \mid c,o)$.
To account for the size differences between human and robot hands, we query the prior with an object point cloud rescaled by the human-to-robot hand scale ratio.
Given sampled modes and wrist initializations, optimization produces robot grasps following $\pi(\bm{Q},\bm{T} \mid \bm{T}_0,c,o)$.
Together, they induce the final grasp distribution $\pi(\bm{Q},\bm{T},c \mid o)$ for cross-mode synthesis.
Finally, we filter grasps in MuJoCo \cite{todorov2012mujoco} and retain all validated grasps across contact modes for each object, rather than only the best one~\cite{yang2026ultradexgrasp}, to preserve dataset diversity for downstream applications.

\textbf{Prior-Guided Optimization Budget Allocation.}
Before optimization, the learned prior predicts $\pi(c \mid o)$ and samples wrist initializations for each contact mode.
For each mode, we draw $4K$ candidates, rank them by approximate diffusion likelihood, keep the top $2K$, and apply farthest-pose selection to retain $K$ diverse human-like initializations.
The retained poses are mapped to the robot hand and slightly perturbed before optimization.
Given a total budget of $B$ optimization attempts, we allocate the mode budget as $B_c = \,\pi(c \mid o) B$ with a maximum per-mode cap.
Each attempt samples one retained wrist initialization from its mode and open-hand joint angles.

\textbf{Grasp Optimization Formulation.}
Given a contact mode $c$ and wrist initialization $\bm{T}_0$, we follow \cite{chen2025bodex} and formulate grasp synthesis as a bi-level optimization, where the outer problem updates $(\bm{Q},\bm{T})$ and the inner problem evaluates force closure.
We instantiate the force-closure energy in Eq.~\ref{eq:grasp_optimization} as $\Phi(g)=\sum_{j=1}^{s}\Phi_j(g)$ over disturbance directions $\{\bm{t}_j\}_{j=1}^{s}$, where each directional residual is the optimal value of the following quadratic program (QP):
\begin{equation}
\Phi_j(g) \triangleq \min_{\bm{F}_j} \left\|\beta \bm{t}_j-\sum_{i=1}^{N_c}\bm{G}_i(g)\bm{f}_{j,i}\right\|^2 \quad \mathrm{s.t.}\ \bm{f}_{j,i}\in\mathcal{F}_i,\ i=1,\dots,N_c;\ \sum_{i=1}^{N_c} f^n_{j,i} \ge \gamma.
\end{equation}
Here, $\bm{F}_j=\{\bm{f}_{j,i}\}_{i=1}^{N_c}$ are contact forces and $\bm{G}_i(g)$ is grasp matrix of contact point $\bm{p}_i$; $\Phi_j$ measures resistance to disturbance $\bm{t}_j$.
Unlike \cite{yang2026ultradexgrasp,lin2026bidexgrasp}, we use a weighted sum of global and per-hand force-closure terms for bimanual synthesis to promote both bimanual stability and individual hand quality.


%% file: sections/results.tex

\section{Results}
\label{sec:result}

\textbf{Evaluation Goals.}
We evaluate whether our object-geometry-aware human prior improves large-scale multi-mode synthesis through better contact mode allocation and wrist initialization, and supports downstream model distillation for real-world deployment across object scales.


\textbf{Implementation and Baselines.}
We construct object scenes from DGN2k \cite{chen2025bodex}, which includes 2.4k meshes.
For each object, we sample multiple tabletop-stable poses and rescale it to 12 target sizes with AABB half-diagonal lengths from 2 cm to 30 cm, yielding 157k scenes spanning diverse geometries, poses, and scales.
Each scene uses $B=40$ optimization attempts in total, with at most $B_c=20$ attempts per mode.
Main results are reported on the Shadow Hand, with LEAP Hand~\cite{shaw2023leap} results in the appendix.
Averaged over contact modes and object scales, HUGS optimizes 28.6 grasps per second on an RTX 4090.
We compare HUGS against heuristic and ablated variants that differ in how contact modes and wrist-pose initializations are allocated.
For the heuristic variants, scalar-scale contact-mode rules are derived from our human grasp dataset, and wrist poses are randomly initialized by convex-hull sampling with palm-facing poses.
Further details are provided in the appendix.
\begingroup
\setlength{\leftmargini}{1.0em}
\setlength{\itemsep}{0pt}
\setlength{\parsep}{0pt}
\setlength{\topsep}{0pt}
\begin{itemize}
    \item \textbf{Heur-Fix.} Uses the fixed \textit{Single-Full} contact mode for every target object and scale, similar to \cite{chen2025bodex}.

    \item \textbf{Heur-Single.} Assigns one hand-designed contact mode to each object scale bin, similar to \cite{yang2026ultradexgrasp}.

    \item \textbf{Heur-Multi.} Assigns multiple hand-designed contact modes to each object scale bin.

    \item \textbf{HUGS-Single.} Uses the HUGS wrist-pose prior, but allows only one mode per scale bin.

    \item \textbf{HUGS.} Our full method uses the human prior to allocate modes and initialize wrists.
\end{itemize}
\endgroup


\subsection{Does Human Prior Predict Contact Modes Better than Scalar-Scale Rules?}

\begin{wrapfigure}{r}{0.36\linewidth}
    \vspace{-1.0em}
    \centering
    \small
    \refstepcounter{table}\label{tab:mode_prediction}
    \textbf{Table~\ref{tab:mode_prediction}.} {Comparison of contact-mode prediction on held-out objects.}


    \setlength{\tabcolsep}{3.5pt}
    \renewcommand{\arraystretch}{1.1}
    \begin{tabular}{lccc}
    \toprule
    Method & KL $\downarrow$ & Prec. $\uparrow$ & Rec. $\uparrow$ \\
    \midrule
    Scale Rules & 0.612 & 0.805 & 0.892 \\
    Human Prior & \textbf{0.300} & \textbf{0.873} & \textbf{0.964} \\
    \bottomrule
    \end{tabular}
    \vspace{-1.2em}
\end{wrapfigure}

We evaluate whether the geometry-aware human prior predicts contact-mode preferences better than rules based only on scalar object scale.
HUGS-Human objects are split 8:2 into train and test sets; scalar-scale baselines map the object diagonal length to mode distributions using training-set statistics.
Table~\ref{tab:mode_prediction} reports KL divergence, soft precision, and soft recall on test set, with metric and baseline details in the appendix.
The human prior achieves lower KL and higher precision and recall, suggesting that contact modes depend on object geometry beyond scalar scale.

\subsection{Does Human-Prior Guidance Improve Efficient Multi-Mode Grasp Synthesis?}

\begin{figure}[t]
    \centering
    \includegraphics[width=\linewidth]{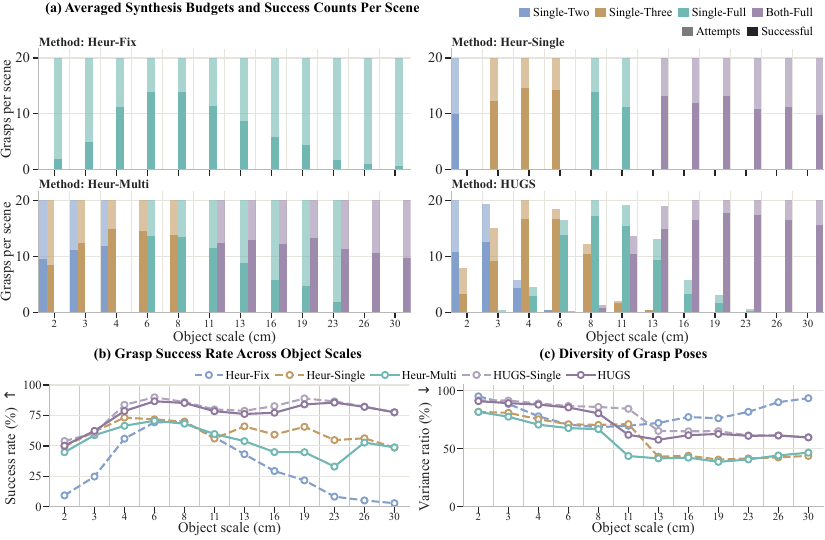}
    \vspace{-1.5em}
    \caption{\textbf{Contact-mode allocation and synthesis success across object scales.} (a) Averaged synthesis budgets and success counts per scene. (b) Overall synthesis success rate across object scales. (c) Pose diversity measured by the explained-variance ratio of the first principal component.}
    \label{fig:synthesis_mode_scale}
    \vspace{-1em}
\end{figure}

We next evaluate how human-prior guidance improves synthesis.
Fig.~\ref{fig:synthesis_mode_scale}(a) shows per-mode budgets (light background bars) and success counts (dark foreground bars), and Fig.~\ref{fig:synthesis_mode_scale}(b) reports overall success across object scales.
The heuristic baselines reveal the limits of scale-only rules.
\textbf{Heur-Fix} works only near the scale suited to \textit{Single-Full}, showing the need for contact-mode adaptation.
\textbf{Heur-Single} is more robust, but one scale-dependent mode cannot cover multiple valid strategies within the same scale.
\textbf{Heur-Multi} adds mode diversity, yet wastes attempts on object-specific mismatches, such as single-hand attempts on 13--23 cm objects.
\textbf{HUGS} instead allocates budget by object geometry, smoothly shifting from \textit{Single-Two} to \textit{Single-Three}, \textit{Single-Full}, and finally \textit{Both-Full} as scale grows.
It also captures object-specific exceptions within the same scale: at 13--19 cm, HUGS makes fewer \textit{Single-Full} attempts but targets the few objects that admit single-hand grasps more accurately, yielding a much higher success rate than Heur-Multi.
Thus, HUGS consistently outperforms all baselines.
Fig.~\ref{fig:teaser} visualizes diverse synthesized grasps, where many adjacent examples show the same object synthesized with different contact modes.
Separately, HUGS-Single outperforms Heur-Single under the same single-mode setting, isolating the benefit of human-prior wrist initialization.

\begin{wrapfigure}{r}{0.37\linewidth}
    \vspace{-1.0em}
    \centering
    \includegraphics[width=\linewidth]{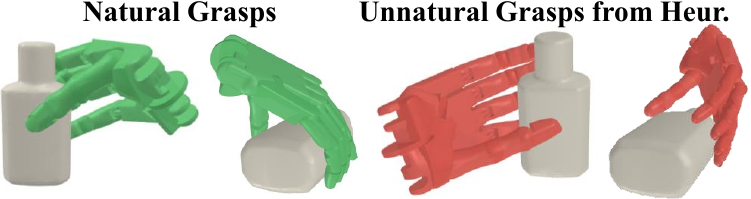}
    \vspace{-1.8em}
    \caption{\small {Heuristic wrist-pose sampling produces unnatural grasps.}}
    \label{fig:unnatural_grasps}
    \vspace{-1.2em}
\end{wrapfigure}
Fig.~\ref{fig:synthesis_mode_scale}(c) compares grasp diversity using the explained-variance ratio of the first principal component \cite{chen2025bodex,chen2025dexonomy}.
HUGS has lower pose dispersion because it concentrates on human-preferred wrist regions.
This is expected: heuristic sampling increases diversity partly via unnatural reversed wrist poses (Fig.~\ref{fig:unnatural_grasps}), which are difficult for humanoid robots to execute.

\subsection{Can Synthesized Grasps Supervise Online Grasp Generators?}

\begin{figure}[b]
    \vspace{-1em}
    \centering
    \includegraphics[width=\linewidth]{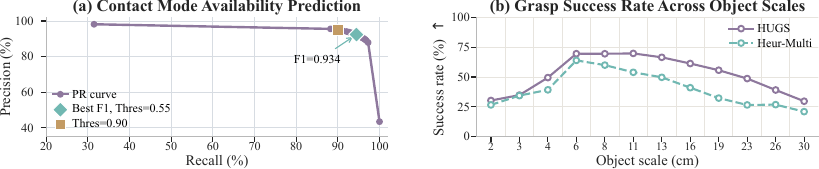}
    \vspace{-1.5em}
    \caption{\textbf{Distilling from synthesized grasps.} (a) Precision-recall curve for contact-mode availability prediction. (b) Grasp success rate, comparing generators trained on HUGS and Heur-Multi data.}
    \label{fig:distill}
    \vspace{-0.5em}
\end{figure}

We further test whether synthesized grasps can supervise learning-based grasp generation.
We train identical lightweight object-conditioned generators on Heur-Multi and HUGS data, both of which contain multiple contact modes for the same object, using an 8:2 DGN2k object split and evaluating on held-out scenes across scales.
The generator predicts binary contact-mode availability and corresponding grasp poses; architecture details are in the appendix.
With HUGS data, contact-mode availability is accurately predicted, reaching a best F1 score of 0.934 in Fig.~\ref{fig:distill}(a).
In simulation, the HUGS-trained generator achieves higher grasp success than the Heur-Multi-trained one (Fig.~\ref{fig:distill}(b)), showing that HUGS produces easier-to-learn training data with better tolerance to generated-grasp errors.
Success remains above 70\% for medium objects (6--13 cm), but drops on very small and large objects, showing that cross-scale, cross-mode grasp generation remains challenging.
Since our focus is grasp synthesis, we leave specialized generator design to future work.

\subsection{Real-World Cross-Scale Grasping Demonstrations}

\begin{figure}[t]
    \centering
    \includegraphics[width=0.95\linewidth]{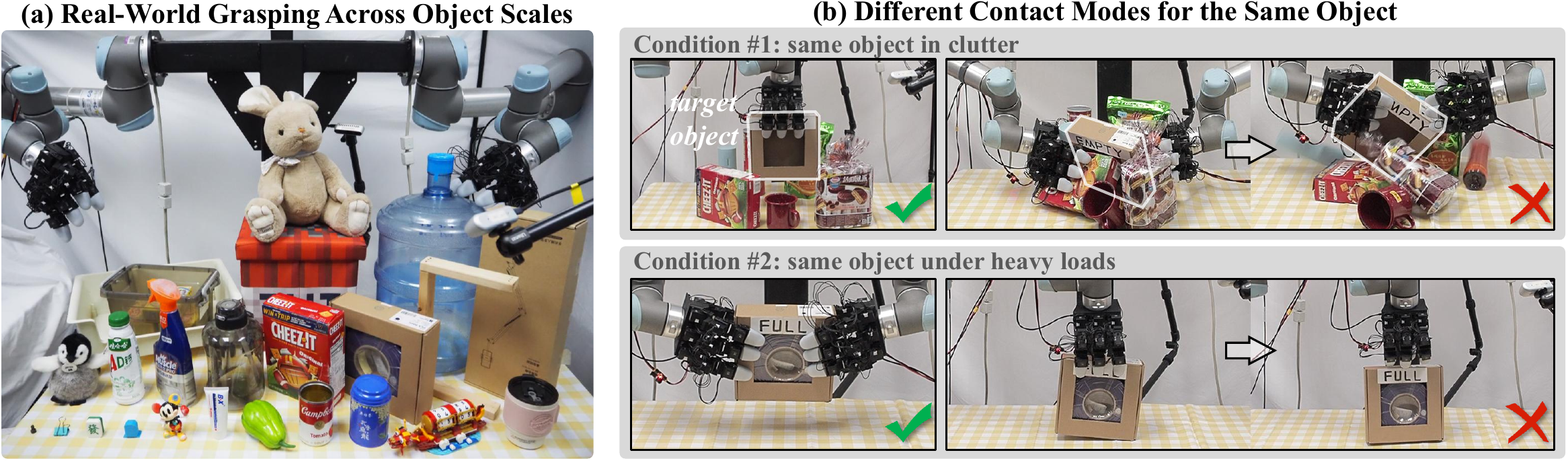}
    \vspace{-0.2em}
    \caption{\textbf{Real-world grasping demonstrations on LEAP Hands.} (a) Cross-scale grasping over diverse objects. (b) Different mode selection under different deployment constraints.}
    \label{fig:real_exp}
    \vspace{-1em}
\end{figure}

We demonstrate real-world grasping with LEAP Hands using a generator trained from synthesized LEAP grasps.
Across 24 objects, from a screw with a 0.5 cm half-diagonal to a bucket with a 30 cm half-diagonal (Fig.~\ref{fig:real_exp}(a)), the generator adaptively selects suitable contact modes and grasp poses.
Fig.~\ref{fig:real_exp}(b) shows why mode diversity matters even for the same object: in clutter, a top-down single-hand grasp remains feasible while side bimanual grasping is blocked; when the box is loaded, bimanual grasping becomes necessary to provide sufficient grasping force.
Thus diverse generated grasps allow deployment-time constraints to select an appropriate grasp.
More grasp cases are shown in Fig.~\ref{fig:teaser} and the appendix; deployment details are also provided in the appendix.
Note that the generator is still preliminary, and perception errors from calibration and depth reconstruction remain important bottlenecks for reliable grasping across object scales.



%% file: sections/conclusion.tex

\section{Limitations and Conclusion}
\label{sec:limitation}
\label{sec:conclusion}

\textbf{Limitations.}
\textit{Human data scale.} HUGS relies on a compact human dataset, so contact-mode and wrist-pose predictions may degrade on far out-of-distribution objects; scaling to large-scale egocentric interaction data is promising but requires further study of mode discovery and annotation.
\textit{Online grasp generation.} Our generator is preliminary and remains less reliable than the offline synthesis pipeline, especially for geometrically irregular objects, motivating research on stronger generative models and geometry representations.
\textit{Contact Diversity.} 
The four contact modes in HUGS capture dominant human grasp strategies but do not exhaustively cover the full spectrum of human grasping behaviors \cite{feix2015grasp}. Extending to denser taxonomies or contact maps is needed for functional grasp synthesis beyond stable lifting.
\textit{Complete grasping systems.} Our real-world demonstrations remain sensitive to calibration, segmentation, and depth reconstruction errors, while robust deployment also requires end-to-end closed-loop execution, tactile adaptation, and collision-aware motion.

\textbf{Conclusion.}
We presented HUGS, a human-prior-guided framework for unified dexterous grasp synthesis across contact modes and object scales.
Instead of directly retargeting human demonstrations, HUGS learns an object-conditioned prior over contact modes and wrist poses to guide robot-specific optimization.
Experiments show that this prior improves contact-mode prediction, synthesis success, and downstream grasp generation, while real-world demonstrations show adaptive mode selection and grasping from tiny screws to large containers.
These results suggest that a human prior learned from compact human grasp data, when used as high-level guidance rather than direct imitation, can support scalable synthesis of diverse and executable dexterous robot grasps across modes and scales.


%% file: sections/appendix.tex

\section{Implementation Details}
\label{app:implementation_details}

\subsection{HUGS-Human Dataset Details}
\label{app:hugs_human_annotation}

\begin{figure}[h]
    \centering
    \includegraphics[width=0.85\linewidth]{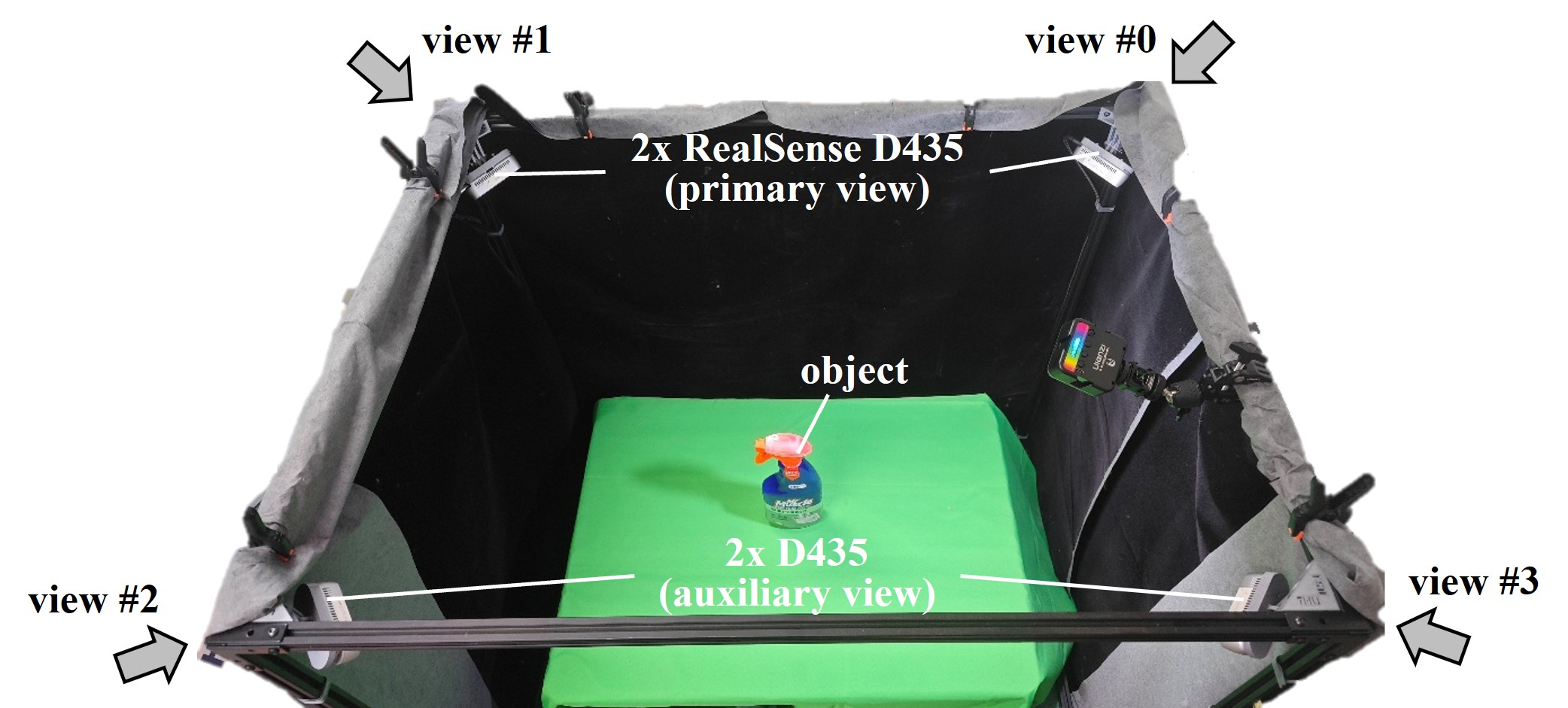}
    \caption{\textbf{Hardware setup for collecting the HUGS-Human dataset.} Four RealSense D435 cameras are used for hand-object mesh reconstruction. Camera views \#0 and \#1 serve as the primary views, while auxiliary views \#2 and \#3 are used to reduce single-view depth ambiguity for each hand.}
    \label{fig:human_grasp_hardware_setup}
\end{figure}

\textbf{Hardware Setup.}
Raw demonstrations are recorded with four RealSense D435 cameras around the tabletop workspace, as shown in Fig.~\ref{fig:human_grasp_hardware_setup}.
All cameras are calibrated for both intrinsic and extrinsic parameters, and the tabletop coordinate frame is also calibrated.
We record RGB-D streams from all four cameras.
The depth images are aligned with the RGB images, and all RGB-D streams from the four camera views are approximately synchronized and captured at a resolution of $1280 \times 720$ at 15 Hz.
The raw recordings cover the complete manipulation process, including approaching the object, lifting it, and placing it back on the table.

\textbf{Annotation Details.}
The dataset annotation includes hand-object mesh reconstruction and discrete contact-mode annotation.
We use camera views \#0 and \#1 as the primary views for reconstructing the right and left hands, respectively, and auxiliary views \#2 and \#3 to reduce single-view depth ambiguity for each hand. View \#1 is used for object mesh reconstruction and pose tracking due to its relatively low occlusion. The depth streams are used only to supervise object reconstruction.
For each raw recording, we use Grounded-SAM2~\cite{ren2024grounded,ravi2024sam2segmentimages} for object segmentation, SAM3D~\cite{chen2025sam} for mesh reconstruction, and FoundationPose~\cite{wen2024foundationpose} for pose estimation. The grasp frame is defined as the frame immediately before clear object motion is detected.
After detecting the grasp frame, we use WiLoR~\cite{potamias2025wilor} to estimate MANO hand poses~\cite{romero2017embodied} from the primary camera views \#0 and \#1. To reduce the depth ambiguity inherent in single-view reconstruction, which can lead to hand-object penetration or separation, we triangulate 3D hand keypoints from the primary and auxiliary views using Direct Linear Transform (DLT). We then solve for the rigid transformation that best aligns the WiLoR detections with the triangulated 3D keypoints.
Finally, the discrete contact mode $c$ is manually annotated according to the number of hands involved in grasping and the number of dominant contacting fingers, following the four modes defined in Sec.~\ref{sec:method}.
For quality control, we manually inspect all annotations and adjust them when necessary, such as correcting the selected grasp frame.
Each data sample includes the object mesh, object pose, hand wrist pose, MANO pose, and contact mode.
We will release both raw recordings and processed annotations.

\begin{figure}[!htbp]
    \centering
    \includegraphics[width=0.95\linewidth]{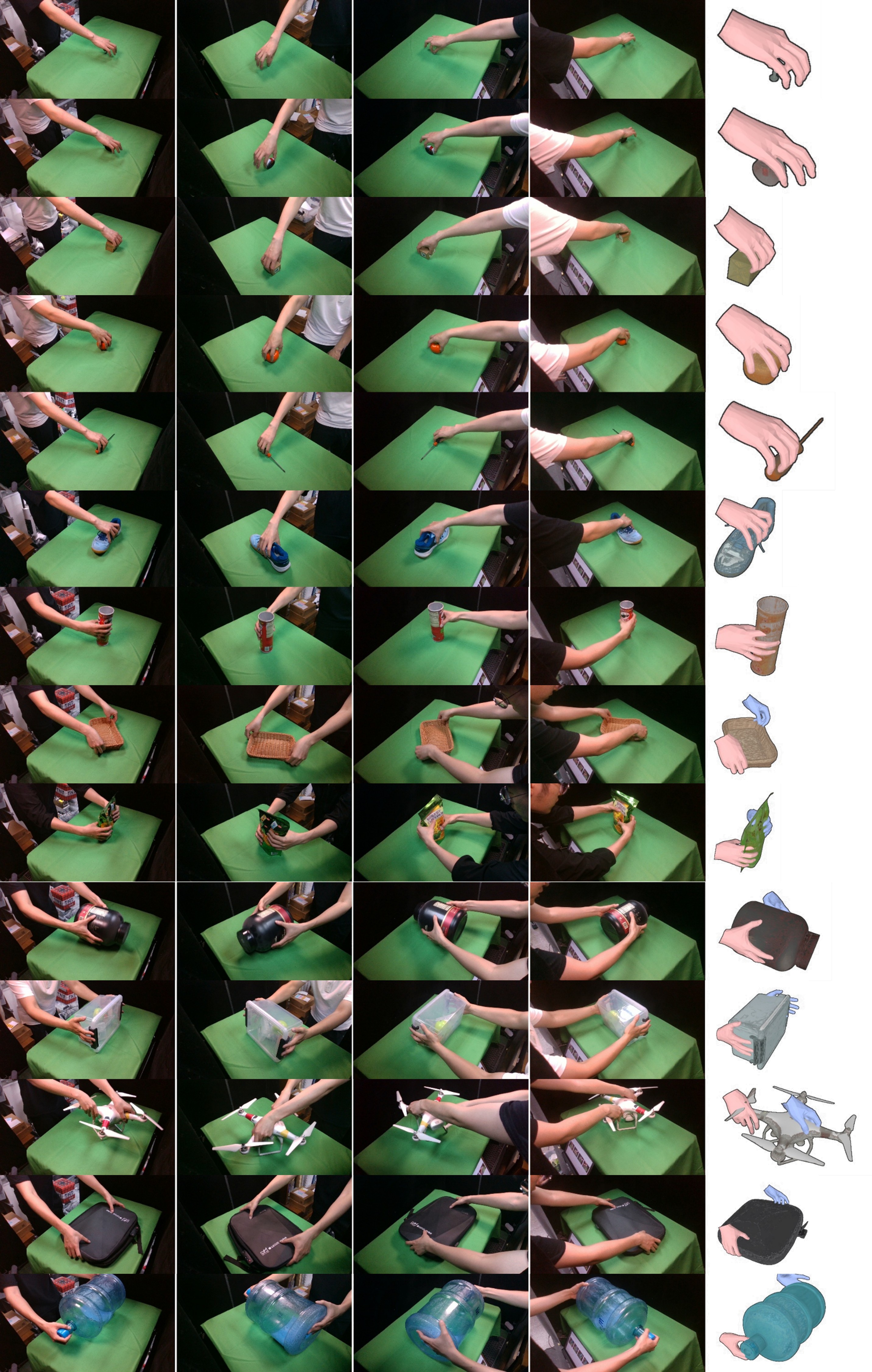}
    \caption{\textbf{Visualization of the HUGS-Human dataset.} Representative samples are shown with object scale increasing from top to bottom. Each sample includes RGB images from all four views, along with the rendered hand-object mesh reconstruction.}
    \label{fig:hugs_human_scale_visualization}
\end{figure}

\begin{figure}[!htbp]
    \centering
    \includegraphics[width=0.95\linewidth]{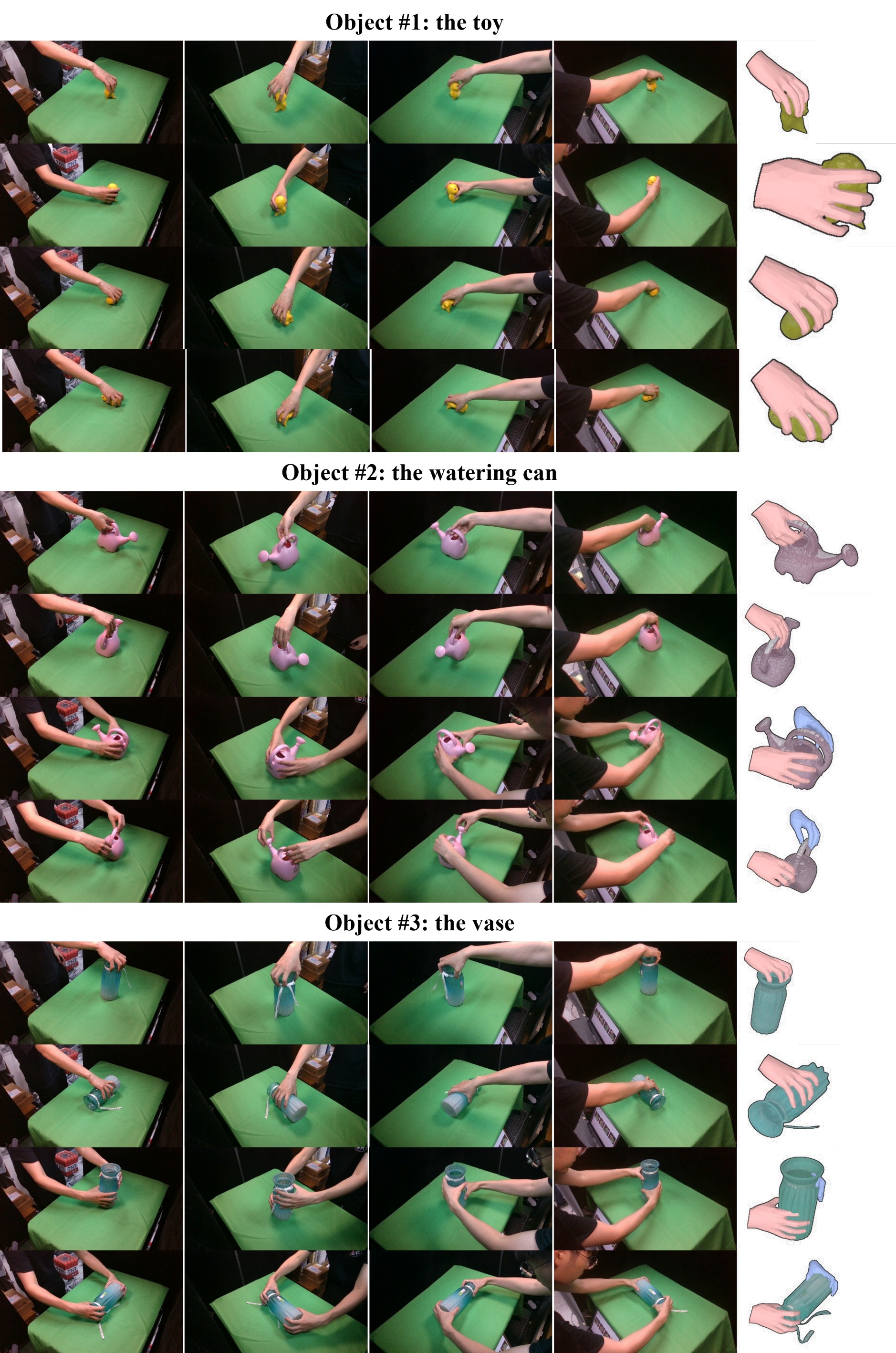}
    \caption{\textbf{Visualization of representative cases of multiple grasps collected with a single object.} For the smaller object \#1, we collect single-handed grasps with varying numbers of fingers and object poses. For the larger objects \#2 and \#3, both single-handed and two-handed grasps are collected.}
    \label{fig:hugs_human_single_object_grasps}
\end{figure}

\textbf{Dataset Visualization.}
The 304 objects in HUGS-Human are collected from common office and laboratory environments.
We visualize the object-scale distribution, per-object grasp count, and contact-mode distribution of HUGS-Human in Fig.~\ref{fig:human_dataset_statistics}.
We further visualize representative samples from the HUGS-Human dataset at different object scales in Fig.~\ref{fig:hugs_human_scale_visualization}, showing both the extracted grasps as reconstructed hand-object meshes and the raw RGB images from all four views.
Finally, we visualize representative cases of multiple grasps collected with a single object in Fig.~\ref{fig:hugs_human_single_object_grasps}, highlighting the object-level diversity of the HUGS-Human dataset.

\begin{figure}[h]
    \centering
    \includegraphics[width=0.6\linewidth]{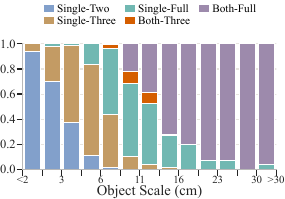}
    \caption{\textbf{Contact-mode distribution when including \textit{Both-Three}.} The \textit{Both-Three} mode accounts for only a negligible fraction of the collected demonstrations.}
    \label{fig:both_three_distribution}
\end{figure}

\textbf{Rare Both-Three Contact Mode.}
At the beginning of data collection, we also planned to include a \textit{Both-Three} contact mode, where two hands are involved but each hand mainly uses sparse three-finger contacts.
After completing the collection, however, we found that only a very small number of objects naturally support this mode, resulting in a negligible fraction of the dataset, as shown in Fig.~\ref{fig:both_three_distribution}.
Incorporating \textit{Both-Three} into the HUGS framework is technically feasible, since our contact-mode prior and optimization pipeline can be extended to additional discrete modes.
Nevertheless, due to the limited number of valid demonstrations, evaluation results for this mode would be statistically unreliable.
We therefore exclude \textit{Both-Three} from the evaluation statistics and focus our analysis on the four dominant contact modes.

\subsection{Human Prior Training Details}
\label{app:human_prior_training}

\begin{figure}[h]
    \centering
    \includegraphics[width=0.8\linewidth]{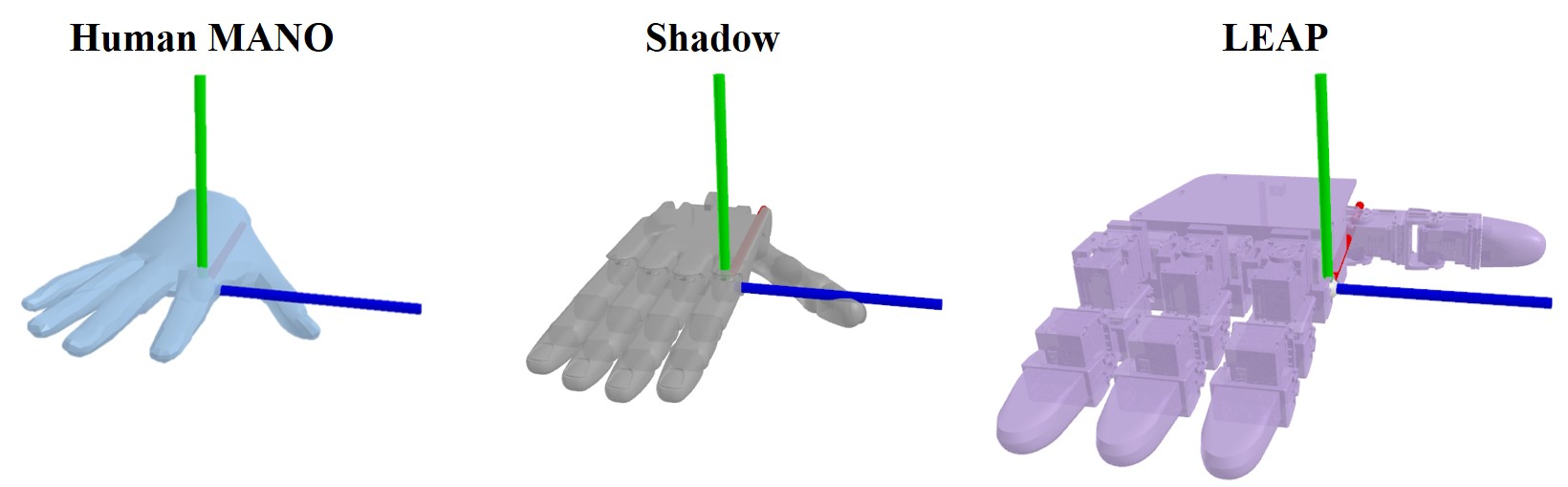}
    \caption{\textbf{Definition of the index MCP frame.} We use the index MCP position as the translation anchor and the dorsal-hand wrist orientation as the rotation reference for the human wrist prior.}
    \label{fig:index_mcp_frame}
\end{figure}

\textbf{Definition of Index MCP Frame.}
To make the human wrist prior less sensitive to embodiment-specific wrist definitions, we define the predicted pose using an index MCP frame, as shown in Fig.~\ref{fig:index_mcp_frame}.
Its translation is the position of the index metacarpophalangeal (MCP) joint, and its rotation follows the dorsal-hand wrist orientation.
This representation provides a stable hand-level pose that can be mapped to different robot hands during synthesis.

\textbf{Architecture and Training.}
In practice, we train the mode and wrist-pose heads separately because contact-mode prediction converges faster and overfits more easily than wrist-pose generation.
Both models take the object point cloud as input.
For each object sample, we uniformly sample 1024 points from the object point cloud and center the point cloud before feeding it to the network.
The centered point cloud is sparsely quantized with MinkowskiEngine using a voxel size of $0.005$.
The point-wise features are aggregated by mean pooling to obtain a 1024-dimensional global object feature.
During training, we apply data augmentation including point-cloud centering, random rotations around the $z$ axis, small rotations around the $x$ and $y$ axes with a maximum angle of $3^\circ$, and uniform scale augmentation in the range $[0.9, 1.1]$.
For scale augmentation, valid hand translation targets are scaled accordingly.
We also add Gaussian noise to point coordinates with standard deviation $0.001$, clipped to three standard deviations, and apply point dropout with a ratio of $0.1$.
The contact-mode prior uses a two-layer MLP head.
We supervise this branch with posed-object-level soft labels.
For multiple human grasp records associated with the same posed object, we compute the occurrence frequency of each contact mode as the target distribution.
It is trained with the soft-label cross-entropy loss
\begin{equation}
    \mathcal{L}_{\mathrm{mode}}
    =
    -\frac{1}{n_{\mathrm{mode}}}\sum_i \sum_k q_{i,k}\log \mathrm{softmax}(\bm{z}_i)_k,
\end{equation}
where \(i\) indexes the samples included in the mode-supervision set, and \(k\) indexes the discrete mode classes.
where $\bm{q}_i$ is the empirical contact-mode distribution computed from all human grasps of the same posed object in the dataset, and $\bm{z}_i$ denotes the predicted logits.
We train this branch from scratch for 100 iterations using AdamW with a batch size of 256.
The learning rate is initialized to $10^{-3}$ and decayed to $10^{-4}$ with a cosine schedule.
The pose prior is a contact-mode-conditioned diffusion model.
We use a learnable 128-dimensional embedding for each contact mode and concatenate it with the 1024-dimensional global object feature to form a 1152-dimensional conditional feature.
Training samples are uniformly drawn at random from all human grasps in HUGS-Human.
The diffusion model generates a normalized 24-dimensional bimanual pose vector $[\bm{R}_r(9), \bm{p}_r(3), \bm{R}_l(9), \bm{p}_l(3)]$.
The translation target $\bm{p}$ represents the index-MCP position, and the rotation target $\bm{R}$ uses the 9D representation of the wrist/palm rotation matrix.
For right-only grasps, the left hand is filled with a fixed placeholder pose.
During training, we apply RMS normalization to the pose vector.
We use a cosine noise schedule with 128 training diffusion steps and optimize the denoiser with a velocity-prediction objective.
The denoiser takes the noised 24D pose, the 1152-dimensional conditional feature, and the timestep embedding as inputs.
It uses hidden layers with Mish activations and outputs a 24D velocity prediction.
We train the denoiser with a SmoothL1 denoising objective between the predicted and target velocities.
At inference time, we use a 10-step DDIM-style deterministic sampler conditioned on the object feature and contact-mode embedding.
We train this branch from scratch for 7500 iterations using AdamW with a batch size of 256.
The learning rate is initialized to $10^{-3}$ and decayed to $10^{-4}$ with a cosine schedule.

\subsection{Grasp Synthesis Details}
\label{app:grasp_synthesis_details}

\textbf{Contact Regions for Each Contact Mode.}
Each contact mode specifies the active fingertip regions used by the robot hand during grasp optimization.
For \textit{Single-Two}, the active contacts are the thumb and index fingertips of one hand.
For \textit{Single-Three}, the active contacts are the thumb, index, and middle fingertips of one hand.
For \textit{Single-Full}, the active contacts are all fingertips of one hand.
For \textit{Both-Full}, the active contacts are all fingertips of both hands.
During grasp optimization, different contact modes only change the expected fingertip contacts involved in the force-closure computation.

\begin{figure}[t]
    \centering
    \includegraphics[width=0.7\linewidth]{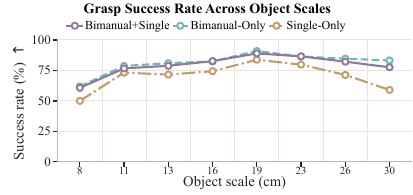}
    \caption{\textbf{Ablation of the bimanual force-closure objective.} We compare using only the per-hand force-closure terms (\textit{Single-Only}), only the global bimanual force-closure term (\textit{Bimanual-Only}), and their combination (\textit{Bimanual+Single}). Without considering bimanual force closure, the grasp success rate drops significantly.}
    \label{fig:ablation_bimanual_fc}
\end{figure}

\begin{figure}[t]
    \centering
    \includegraphics[width=0.8\linewidth]{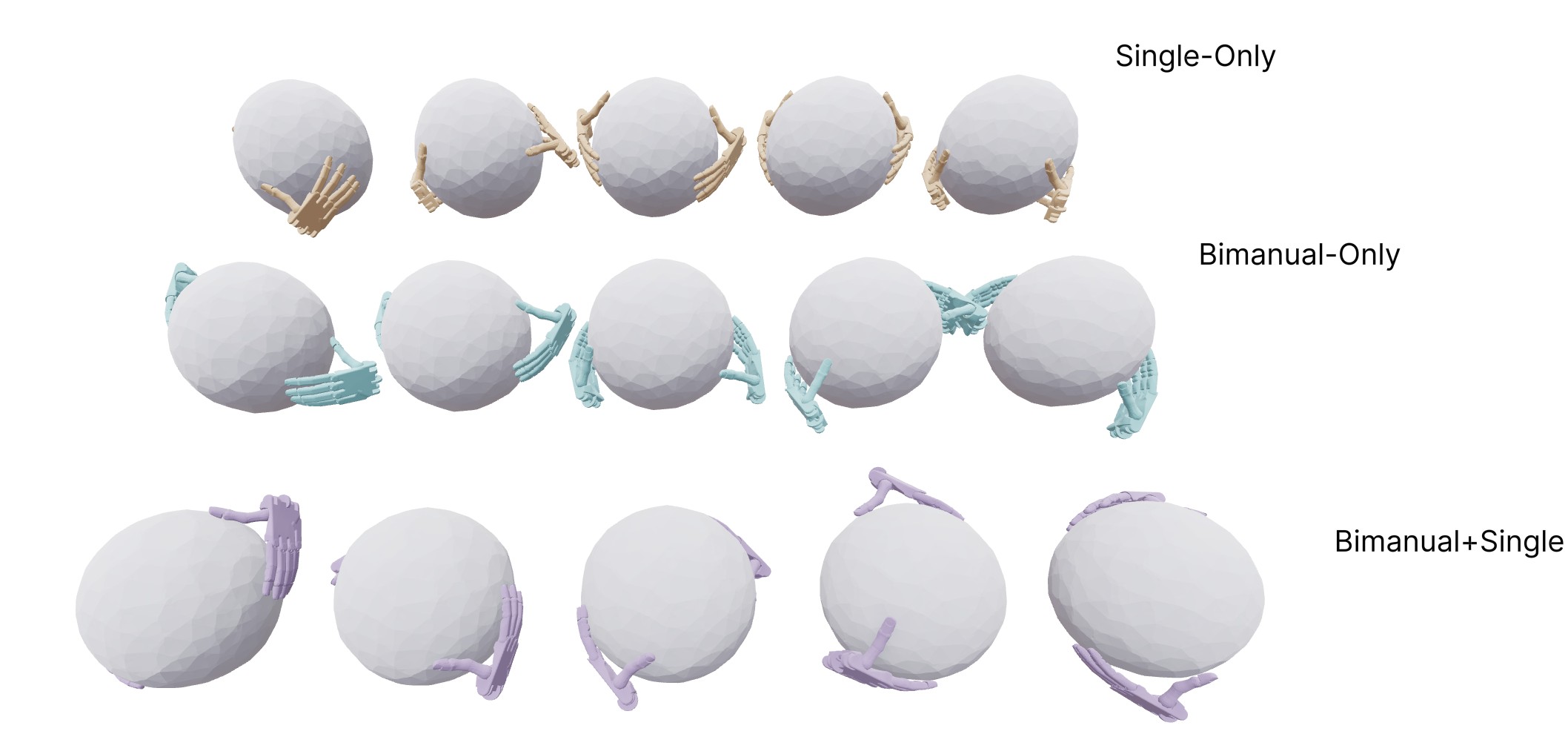}
    \caption{\textbf{Qualitative examples of bimanual force-closure ablation.} Using only the global bimanual term (\textit{Bimanual-Only}) can produce degenerate contacts, where each hand touches the object only weakly or at fingertip extremities. Combining global and per-hand terms (\textit{Bimanual+Single}) yields more coordinated bimanual grasps while maintaining better individual hand contact quality.}
    \label{fig:ablation_bimanual_fc_cases}
\end{figure}

\textbf{Bimanual Optimization Implementation.}
To adapt bimanual grasp optimization to the frameworks of \cite{chen2025bodex,sundaralingam2023curobo}, we model the two hands as a single composite robot by adding dummy joints.
The dummy joints include three translational joints and three rotational joints, which parameterize the floating wrist pose of each hand.
In practice, very large objects may lie near or beyond the boundary of the human grasp dataset distribution, where training samples are relatively sparse. As a result, the predicted global hand pose can be slightly less accurate, leading to noticeable hand-object penetration in some cases.
To make the initial states more suitable for subsequent optimization, we offset each hand by 10\,cm along its palm-outward direction when constructing the initial robot wrist poses, as shown in Fig. \ref{fig:prior_to_init_to_grasp}.
After sampling wrist poses from the human prior, we first solve batched inverse kinematics to obtain the corresponding dummy joint angles.
These IK solutions are then used as the initialization for the subsequent grasp optimization.
Unlike the default single-hand configuration in \cite{chen2025bodex}, bimanual optimization first optimizes the in-contact grasp pose, and then optimizes the non-contact pregrasp pose in a final stage by reducing the hand-to-object point distance threshold by 1\,cm.
This ordering allows the grasp pose to be closer to the object center.
In addition, we assign a larger step size to the dummy joint (base pose) variables, which decays over optimization iterations, allowing the wrist poses to adjust more aggressively in early iterations while converging stably.
For the force-closure objective, we use a sum of a global bimanual term and per-hand terms: $\Phi_{\mathrm{bi}}(g) = \Phi_{\mathrm{global}}(g) + \Phi_{\mathrm{left}}(g) + \Phi_{\mathrm{right}}(g)$, where $\Phi_{\mathrm{global}}$ evaluates force closure over contacts from both hands jointly, and $\Phi_{\mathrm{left}}$, $\Phi_{\mathrm{right}}$ evaluate each hand independently.
Fig.~\ref{fig:ablation_bimanual_fc} shows that omitting the global bimanual force-closure term substantially reduces grasp success, indicating that per-hand force closure alone is insufficient for stable bimanual grasping.
Although \textit{Bimanual-Only} achieves a success rate similar to the combined objective, Fig.~\ref{fig:ablation_bimanual_fc_cases} shows that using only the global bimanual term can produce degenerate contacts, where each hand touches the object only weakly or at fingertip extremities.
Therefore, the global term is needed for overall bimanual stability, while the per-hand terms improve the contact quality of each individual hand.

\textbf{Human-to-Robot Scale Ratio.}
Before querying the human prior, we rescale the object point cloud according to the human-to-robot hand scale ratio.
The motivation is to match the relative object size perceived by the robot hand to that in the human demonstrations, since different hand embodiments have different physical sizes as illustrated in Fig.~\ref{fig:index_mcp_frame}.
For example, the LEAP Hand is larger than a human hand, so an object grasped by a human appears relatively smaller to LEAP.
Therefore, when querying the human prior for LEAP synthesis, we downscale the object point cloud before prediction.
We use a scale ratio of $1.0$ for the Shadow Hand and $1.4$ for the LEAP Hand.

\textbf{MuJoCo Simulation.}
We filter optimized grasps in MuJoCo simulation by lifting the object and checking three criteria for success.
First, the object must be lifted to a height of 0.1\,m.
Second, the object pose must remain stable throughout the lift, with position deviation below 5\,cm and rotation deviation below 15\textdegree{}.
Third, the grasp must be collision-free. A grasp is disqualified if the hand is in self-collision or intersects the object at the start of the simulation, or if self-collision occurs at any point during the grasping process.
For physical simulation across object scales, we assign a scale-dependent effective density to each object.
To avoid cubic mass growth under geometric scaling, we use $d(s) = d_0 (s / s_0)^{-2}$, where $d_0 = 700\,\text{kg/m}^3$ and $s_0 = 0.06\,\text{m}$.
This formulation makes the resulting object mass scale approximately linearly with $s$.
The resulting median object masses are 0.032\,kg at scale 0.02, 0.096\,kg at scale 0.06, 0.160\,kg at scale 0.1, and 0.339\,kg at scale 0.3.

\subsection{Grasp Generator Details}
\label{app:grasp_generator}

\textbf{Architecture.}
The robot grasp prior uses the same sparse point-cloud encoder as the human prior.
The input partial object point cloud is fused from three camera views.
We sample 1024 points from the partial object point cloud, apply MinkowskiEngine sparse quantization, and feed the quantized point cloud into MinkUNet.
Mean pooling over point-wise features produces a 1024-dimensional object feature.
The model contains a contact-mode availability head and a mode-conditioned robot pose head.
The availability head predicts independent binary availability scores for the four contact modes.
Its supervision is computed from the contact-mode records of the same scene in the synthetic robot dataset, allowing multiple contact modes to be available simultaneously.
The pose-head architecture follows \cite{chen2025dexonomy}.
It conditions on the concatenation of the object feature and a learnable 128-dimensional contact-mode embedding.
It first uses a conditional diffusion model to generate the final grasp-frame 24-dimensional bimanual wrist pose $[\bm{R}_r, \bm{p}_r, \bm{R}_l, \bm{p}_l]$, where rotations use the 9D rotation-matrix representation.
The diffusion model uses RMS-normalized poses, 128 training diffusion steps, a cosine noise schedule, and a velocity-prediction objective, and is trained with a SmoothL1 denoising loss.
To recover the full robot trajectory, we further use two MLPs for the left and right hands.
Conditioned on the final wrist pose, these MLPs regress wrist translations and rotations for the pregrasp and grasp stages, as well as finger joint positions for the pregrasp, grasp, and squeeze stages.

\textbf{Training.}
The overall training objective consists of five terms: contact-mode availability BCE loss, final-wrist diffusion denoising loss, trajectory translation loss, trajectory rotation loss, and joint-position loss.
We use equal weights for all five terms in the current configuration and train the entire network jointly in a single stage for 50000 iterations.
We use AdamW with a batch size of 256.
The learning rate is initialized to $10^{-3}$ and decayed to $10^{-4}$ with a cosine schedule.

\textbf{Sampling.}
At inference time, the availability head first predicts independent contact-mode scores for the input partial point cloud.
We select feasible contact modes using a score threshold of $0.9$.
According to the precision-recall curve in Fig.~\ref{fig:distill}(a), this threshold yields approximately $95\%$ precision and $90\%$ recall for the Shadow Hand.
For each selected contact mode, we sample the final wrist pose with the DDIM deterministic sampler conditioned on the object feature and contact-mode embedding, followed by RMS de-normalization.
The trajectory MLPs then predict the pregrasp and grasp wrist poses, as well as the pregrasp, grasp, and squeeze joint positions, conditioned on the sampled final wrist pose.
For each feasible contact mode, we sample 100 grasp poses and select the top 10 candidates according to the approximate diffusion probability for simulation evaluation.
The resulting candidates may be further filtered by reachability, collision checking, and task-specific deployment constraints for downstream applications.

\subsection{Real-World Deployment Details}
\label{app:real_world_deployment}

\begin{figure}[t]
    \centering
    \begin{minipage}{0.48\linewidth}
        \centering
        \includegraphics[width=0.99\linewidth]{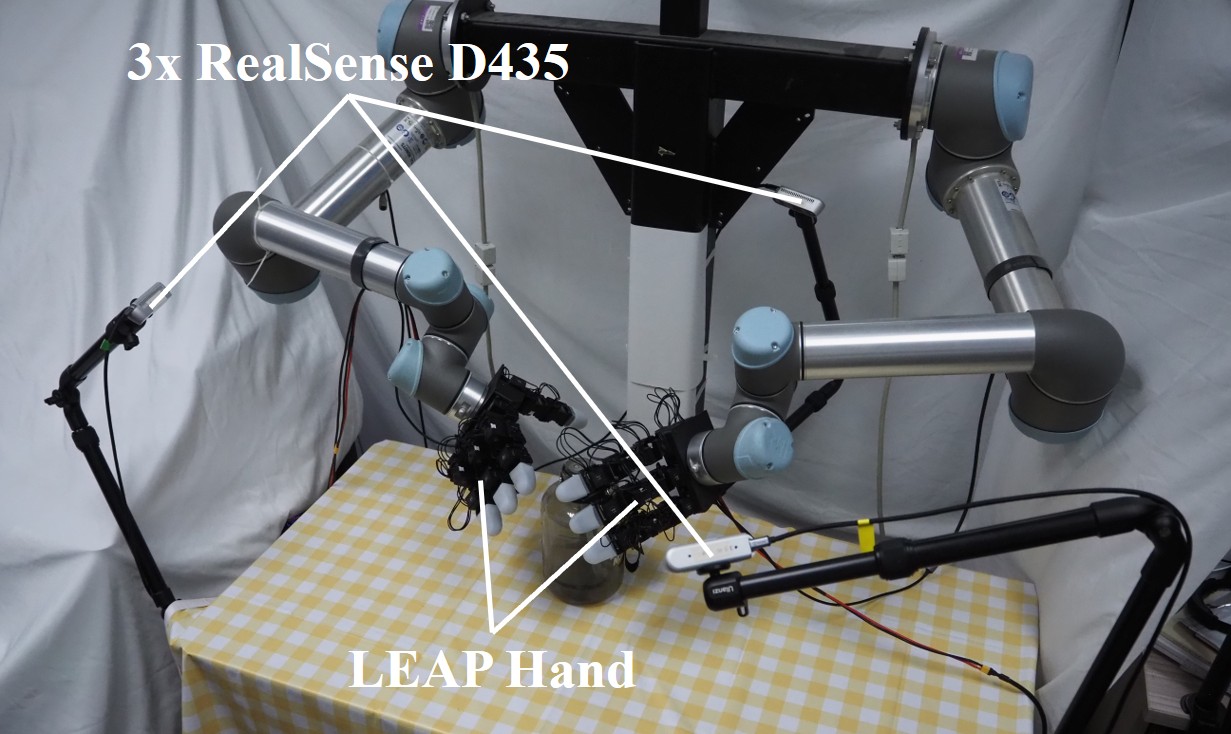}
        \vspace{0.2em}
        \centerline{\small (a) Real-world hardware setup}
    \end{minipage}
    \hfill
    \begin{minipage}{0.48\linewidth}
        \centering
        \includegraphics[width=0.99\linewidth]{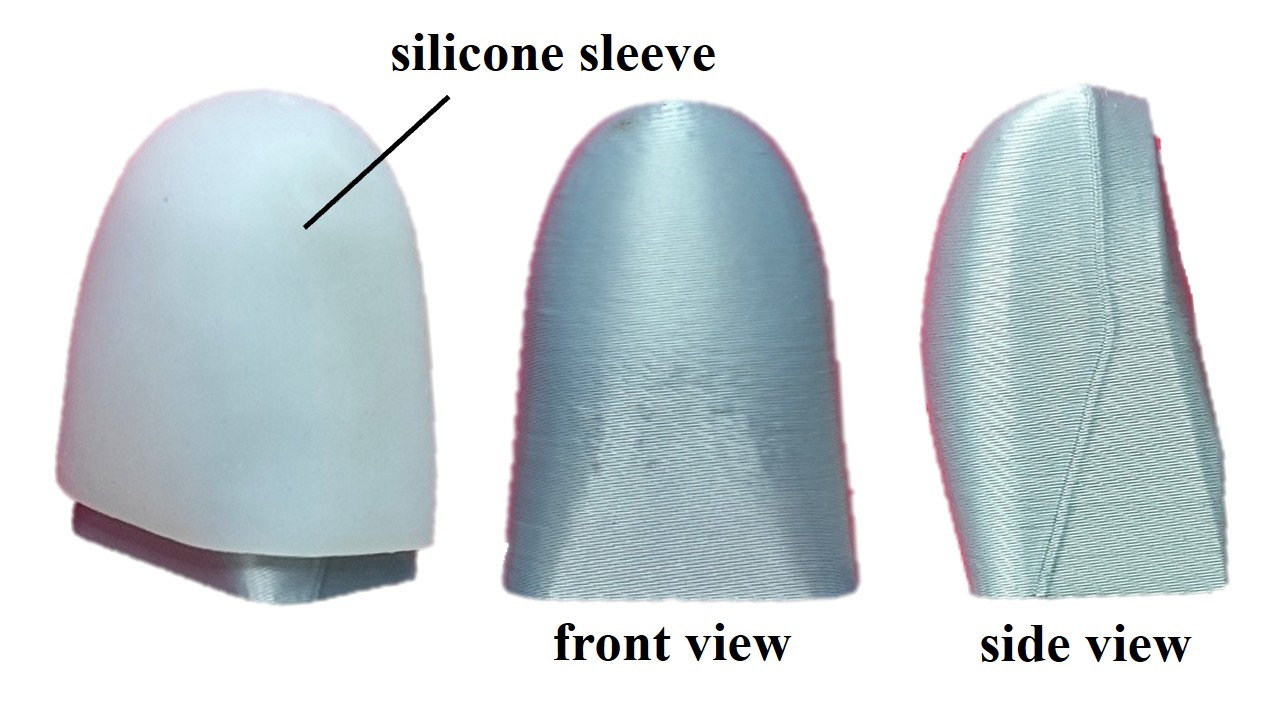}
        \vspace{0.2em}
        \centerline{\small (b) Customized LEAP fingertip}
    \end{minipage}
    \caption{\textbf{Real-world hardware setup and customized LEAP fingertip.} Our hardware setup consists of three RealSense D435 cameras arranged approximately 120$^\circ$ apart for object point-cloud reconstruction. Compared with the original LEAP fingertip, our customized fingertip has a smoother geometry and is covered with a silicone sleeve to improve contact stability.}
    \label{fig:real_world_hardware_setup}
\end{figure}

\textbf{Deployment Pipeline.}
The real-world hardware setup is shown in Fig.~\ref{fig:real_world_hardware_setup}. We deploy the system on a dual-arm UR5 platform with two LEAP Hands equipped with customized fingertips.
During deployment, three RealSense D435 cameras capture RGB-D observations. All cameras are calibrated for inter-camera extrinsics, and we further perform hand-eye calibration between the robot and the primary camera.
RGB images are segmented with Grounded-SAM2~\cite{ren2024grounded,ravi2024sam2segmentimages}, depth maps are refined with Lingbot-depth~\cite{tan2026maskeddepth}, and the fused object point cloud is fed to the generator.
Collision-free approach motions are planned using cuRobo~\cite{sundaralingam2023curobo}.
For real-world execution, we sample 25 grasps for each feasible contact mode and manually select three candidates for robot execution.
Snapshots of the executed real-world grasps on objects \#1--\#23 are shown in Fig.~\ref{fig:real_grasp_1} and Fig.~\ref{fig:real_grasp_2}, while object \#24 is used to demonstrate the significance of contact-mode diversity (Fig.~\ref{fig:real_exp} (b)).
As discussed in Sec.~\ref{sec:limitation}, real-world generation is affected by calibration errors, depth reconstruction noise, segmentation quality, and other deployment-specific factors.
Moreover, our online generator is still preliminary and remains less reliable than the offline synthesis pipeline.
Therefore, the real-world experiments are intended to demonstrate the deployment potential of HUGS in physical scenes, rather than to provide a quantitative evaluation of grasp success rate.
The typical failure modes are analyzed in Appendix~\ref{app:real_world_failure_cases}.
Building a fully autonomous and robust real-world grasping system remains an important direction for future work.

\section{Evaluation Details}
\label{app:evaluation_details}

\subsection{Metrics Details}

\textbf{Human Contact-Mode Distribution Prediction.}
This section provides the metric definitions for the contact-mode distribution prediction experiment in Sec.~\ref{sec:result}.
Given a posed object, the task is to predict the empirical distribution of human-preferred contact modes rather than a single contact-mode label.
Let $\bm{q}$ denote the empirical contact-mode distribution and $\bm{p}$ denote the predicted distribution over contact modes.
We measure the discrepancy between the two distributions using
\begin{equation}
    D_{\mathrm{KL}}(\bm{q}\|\bm{p})=\sum_t q_t \log \frac{q_t}{p_t}.
\end{equation}
We also report soft precision,
\begin{equation}
    \mathrm{SoftPrec}(\bm{p},\bm{q})=\sum_{t\in\mathcal{P}(\bm{q})} p_t,
    \quad
    \mathcal{P}(\bm{q})=\{t\mid q_t>0\},
\end{equation}
which measures how much predicted probability mass falls on ground-truth positive contact modes.
Finally, we report soft recall,
\begin{equation}
    \mathrm{SoftRec}(\bm{p},\bm{q})=\sum_{t\in\mathrm{Top}_{|\mathcal{P}(\bm{q})|}(\bm{p})} q_t,
\end{equation}
which measures how much ground-truth probability mass is covered by the top predicted modes, with the number of selected modes set to \(|\mathcal{P}(\bm{q})|\), i.e., the number of ground-truth positive contact modes.
Lower KL and higher soft precision/recall indicate better contact-mode prediction.

\textbf{Synthesis Success Rate.}
For a given evaluation group, the synthesis success rate is defined as the fraction of optimization attempts that pass physical validation in simulation, i.e., $\mathrm{SuccessRate}=N_{\mathrm{success}}/N_{\mathrm{attempt}}$,
where $N_{\mathrm{attempt}}$ is the number of grasp optimization attempts and $N_{\mathrm{success}}$ is the number of physically successful grasps.

\textbf{Success Counts per Scene.}
For each object scene, the success count is the number of successful grasps obtained from all optimization attempts allocated to that scene.
When reporting success counts per scene across an object-scale bin or a method, we average this count over all scenes in the corresponding group.

\textbf{Pose Diversity.}
We measure grasp-pose diversity using the explained-variance ratio of the first principal component, following the PCA-based diversity metric in prior work~\cite{chen2025bodex,chen2025dexonomy}.
For each synthesized grasp, we use the optimized wrist pose and hand joint angles as the pose feature, including wrist translation, wrist rotation, and finger joint positions.
For each object scene, we perform PCA over all validated grasp poses of that scene and record the explained-variance ratio of the first principal component.
We then average this value over all scenes in the corresponding evaluation group.
This differs from \cite{chen2025bodex,chen2025dexonomy}, which compute PCA after pooling grasps across scenes.
We adopt the scene-wise formulation because our focus is the diversity of valid grasp poses for the same scene.
A lower first-component explained-variance ratio indicates that successful grasps are less dominated by a single pose direction and are therefore less concentrated along one major mode.
In our analysis, however, large dispersion can also arise from unnatural wrist poses produced by heuristic sampling; thus the diversity metric should be interpreted together with grasp naturalness and qualitative examples.

\textbf{Robot Contact-Mode Availability Prediction.}
For the robot grasp generator, contact-mode availability is evaluated as a multi-label binary prediction problem.
For each scene and contact mode, the ground-truth label is positive if the synthesized dataset contains at least one validated grasp of that mode for the scene, and negative otherwise.
Given predicted availability scores, we sweep the decision threshold to compute precision and recall:
$\mathrm{Precision}=\mathrm{TP}/(\mathrm{TP}+\mathrm{FP})$ and $\mathrm{Recall}=\mathrm{TP}/(\mathrm{TP}+\mathrm{FN})$.
The F1 score is defined as $\mathrm{F1}=2\,\mathrm{Precision}\,\mathrm{Recall}/(\mathrm{Precision}+\mathrm{Recall})$.
We report the precision-recall curve and the best F1 score over thresholds.

\subsection{Baseline Details}
\label{app:baseline_details}

\textbf{Scalar-Scale Rules.}
The scale-rule baseline is derived from statistics of HUGS-Human.
Here, object scale refers to the half-diagonal length of the object's axis-aligned bounding box (AABB).
We first group posed objects in the training split into scale bins according to their object size.
For each scale bin, we aggregate all human grasps associated with posed objects in that bin and compute the empirical contact-mode distribution.
At test time, a posed object is assigned to the corresponding scale bin based only on its scalar object size, and the precomputed bin-level contact-mode distribution is used as the prediction.
This baseline therefore captures scale-dependent mode preferences but ignores detailed object geometry.
In the contact-mode distribution prediction experiment in Sec.~\ref{sec:result}, we evaluate the continuous contact-mode distributions directly.
For grasp synthesis, however, the scalar-scale rules are used only to decide whether each contact mode should be considered.
All contact modes selected by the rule are assigned the same optimization budget.

\textbf{Heuristic Wrist-Pose Sampling.}
For heuristic baselines, we use convex-hull wrist sampling to initialize optimization.
For each object scene, we first construct an expanded convex hull of the object.
After applying the object scale in the scene, we expand the hull along the hull-vertex normals.
We then uniformly sample candidate points on the expanded hull surface and transform both the sampled points and face normals into the world frame.
Each candidate point defines a palm seed.
The translation is set to the sampled point, and the approach axis is set opposite to the outward normal so that the palm faces the object.
Candidate seeds are filtered by collision-free checks.
For bimanual grasping, we first randomly sample a right-hand surface candidate.
We then compute the mean $xy$ position of all valid candidates in the current scene as the pairing center, mirror the right-hand $xy$ position about this center, and keep the right-hand $z$ coordinate unchanged.
In the left-hand candidate set, we select the real surface candidate whose translation is closest to this mirrored target.
This procedure only pairs translations; the left-hand rotation is still constructed from the surface normal of the selected left-hand candidate, rather than by mirroring the right-hand orientation.
Finally, we add translational and rotational perturbations to the sampled poses and convert them into initial poses.

\begin{figure}[t]
    \centering
    \includegraphics[width=\linewidth]{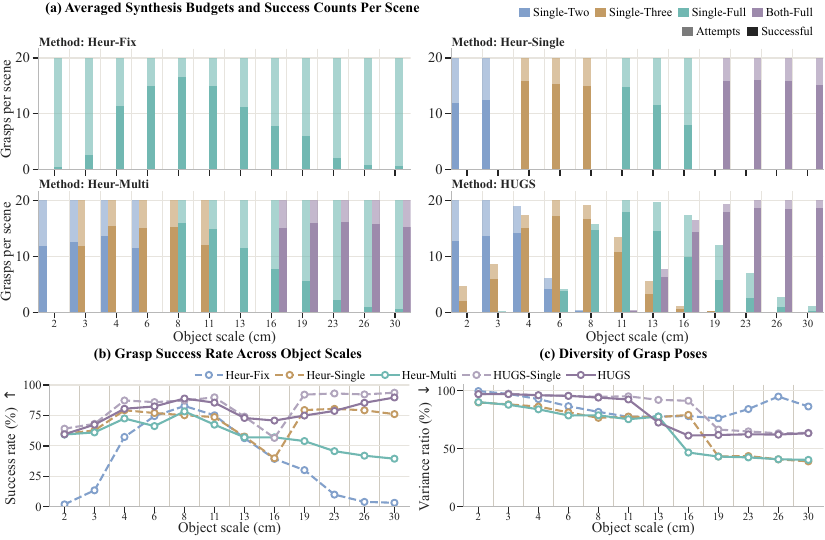}
    \caption{\textbf{Contact-mode allocation and synthesis success on the LEAP Hand.} (a) Averaged synthesis budgets and success counts per scene. (b) Overall synthesis success rate across object scales. (c) Pose diversity measured by the explained-variance ratio of the first principal component.}
    \label{fig:synthesis_mode_scale_leap}
\end{figure}

\section{Additional Results}
\label{app:additional_results}

\subsection{LEAP Hand Simulation Results}
\label{app:leap_results}

We provide additional LEAP Hand simulation results using the same synthesis protocol as the Shadow Hand experiments, except that we use a human-to-robot scale ratio of $1.4$ when querying the human prior.
Since the LEAP Hand is larger than a human hand, an object grasped by a human appears relatively smaller to LEAP; therefore, we downscale the object point cloud before prediction.

\textbf{LEAP Synthesis Results.}
The LEAP Hand results in Fig.~\ref{fig:synthesis_mode_scale_leap} show trends consistent with the Shadow Hand results in Sec.~\ref{sec:result}.
Across object scales, HUGS adaptively reallocates synthesis budget across contact modes instead of relying on fixed scalar-scale rules.
This object-conditioned allocation improves synthesis success over heuristic baselines while preserving multi-mode grasp coverage.
The diversity trend is also consistent with the Shadow Hand experiments: HUGS concentrates optimization around human-preferred wrist regions, whereas heuristic sampling can introduce larger pose dispersion through less natural wrist initializations.
These results suggest that the learned human prior is not tied to a single robot hand and can transfer to a different dexterous hand, even when its physical size differs from that of a human hand.

\begin{figure}[t]
    \centering
    \includegraphics[width=\linewidth]{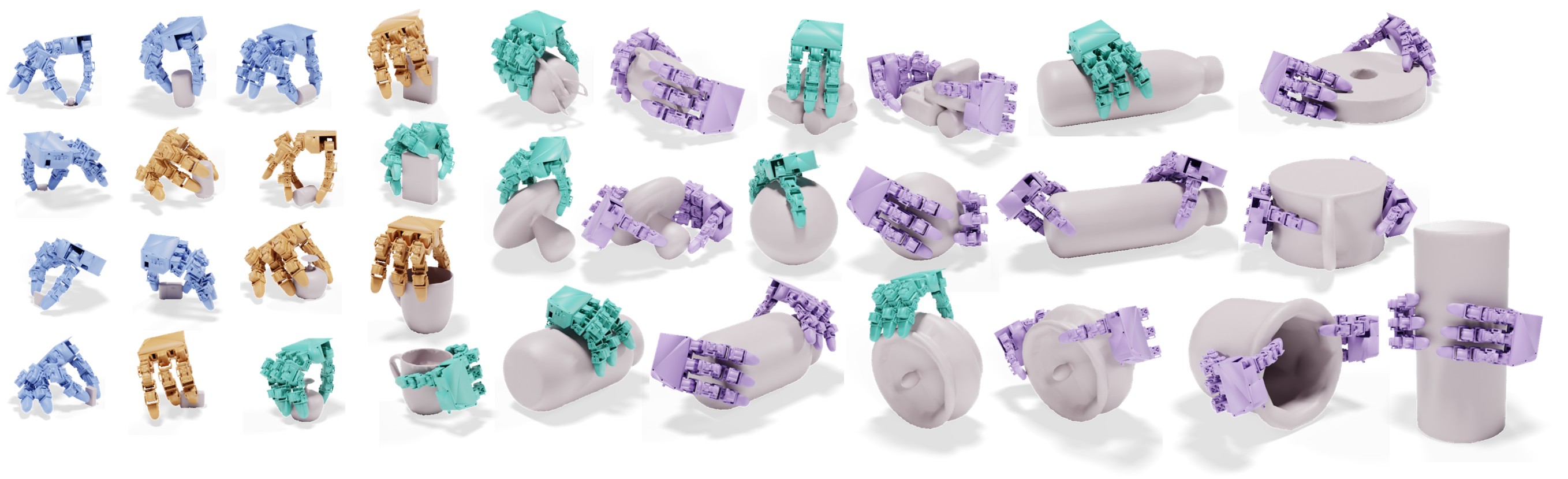}
    \caption{\textbf{Synthesized LEAP Hand grasps across object scales and contact modes.}}
    \label{fig:app:leap_synthetic_grasps}
\end{figure}

\textbf{Qualitative Synthetic LEAP Grasps.}
Fig.~\ref{fig:app:leap_synthetic_grasps} shows qualitative examples of HUGS-synthesized grasps on the LEAP Hand.
The examples span different object scales and contact modes, from sparse single-hand grasps on small objects to full-hand and bimanual grasps on larger objects.
They also include cases where different contact modes are synthesized for the same object.

\begin{figure}[t]
    \centering
    \includegraphics[width=\linewidth]{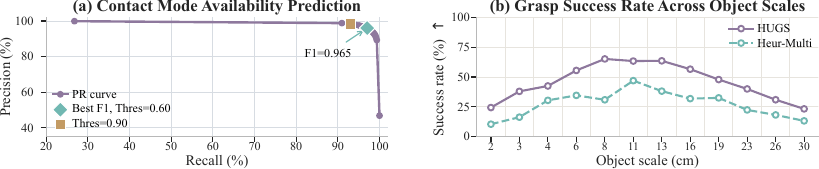}
    \caption{\textbf{Distilling from synthesized LEAP grasps.} (a) Precision-recall curve for contact-mode availability prediction. (b) Grasp success rate, comparing generators trained on HUGS and Heur-Multi LEAP grasp data.}
    \label{fig:distill_leap}
\end{figure}

\textbf{LEAP Distillation Results.}
Fig.~\ref{fig:distill_leap} further evaluates distillation from synthesized LEAP grasps, following the same protocol as the Shadow Hand generator experiment in Fig.~\ref{fig:distill}.
The results are consistent with the Shadow Hand setting: the generator trained on HUGS data predicts contact-mode availability and produces more successful grasps than the generator trained on heuristic data.
This indicates that the synthesized HUGS grasps remain useful for supervising online grasp generators after transferring the synthesis pipeline to the LEAP Hand.

\subsection{More Visualization}

\begin{figure}[t]
    \centering
    \includegraphics[width=0.9\linewidth]{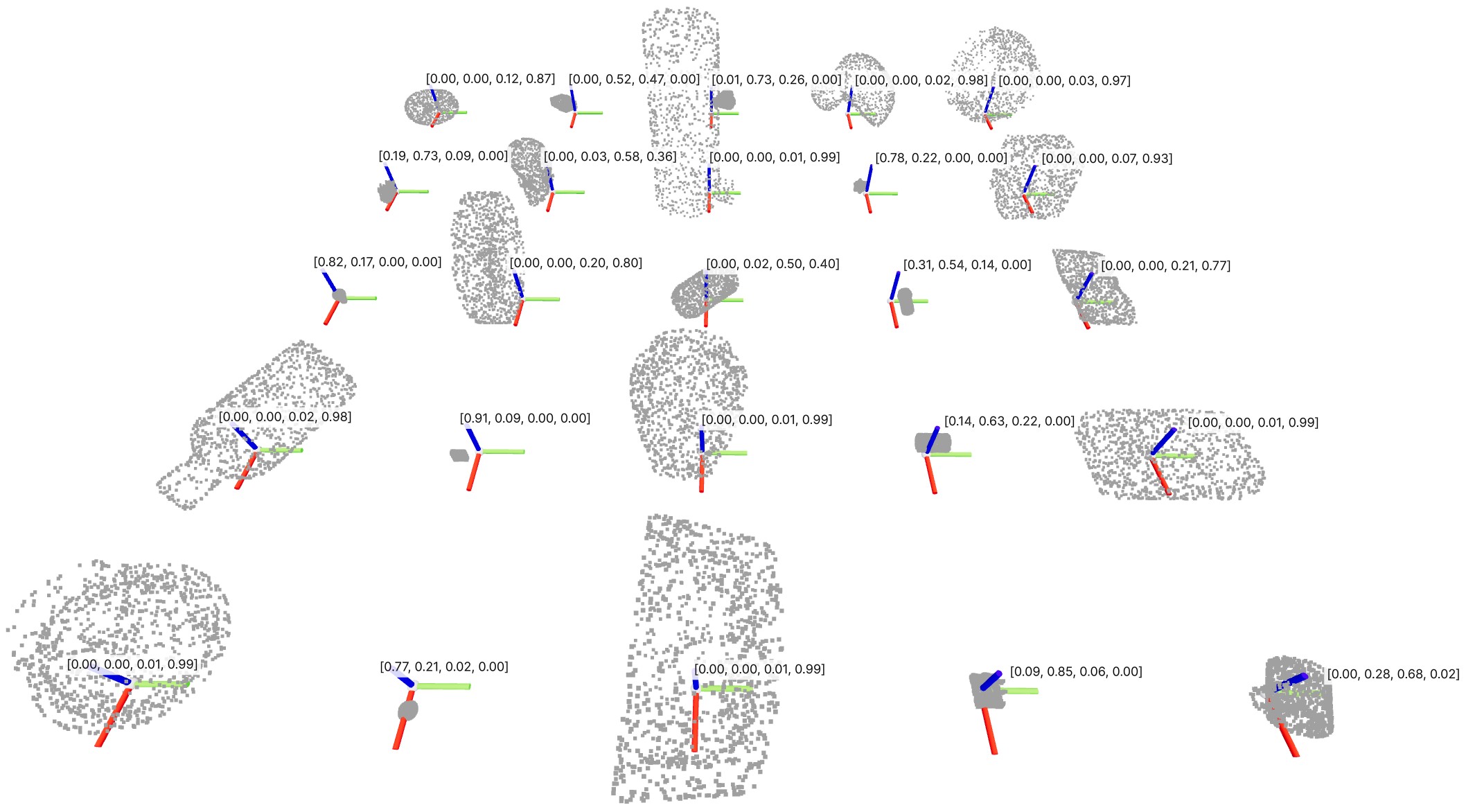}
    \caption{\textbf{Predicted contact-mode distributions from the human prior.} Each example shows an object point cloud together with the predicted probability vector over contact modes, ordered as [\textit{Single-Two}, \textit{Single-Three}, \textit{Single-Full}, \textit{Both-Full}]. The coordinate axes have a length of 0.1\,m.}
    \label{fig:human_prior_contact_mode}
\end{figure}

\begin{figure}[t]
    \centering
    \includegraphics[width=0.9\linewidth]{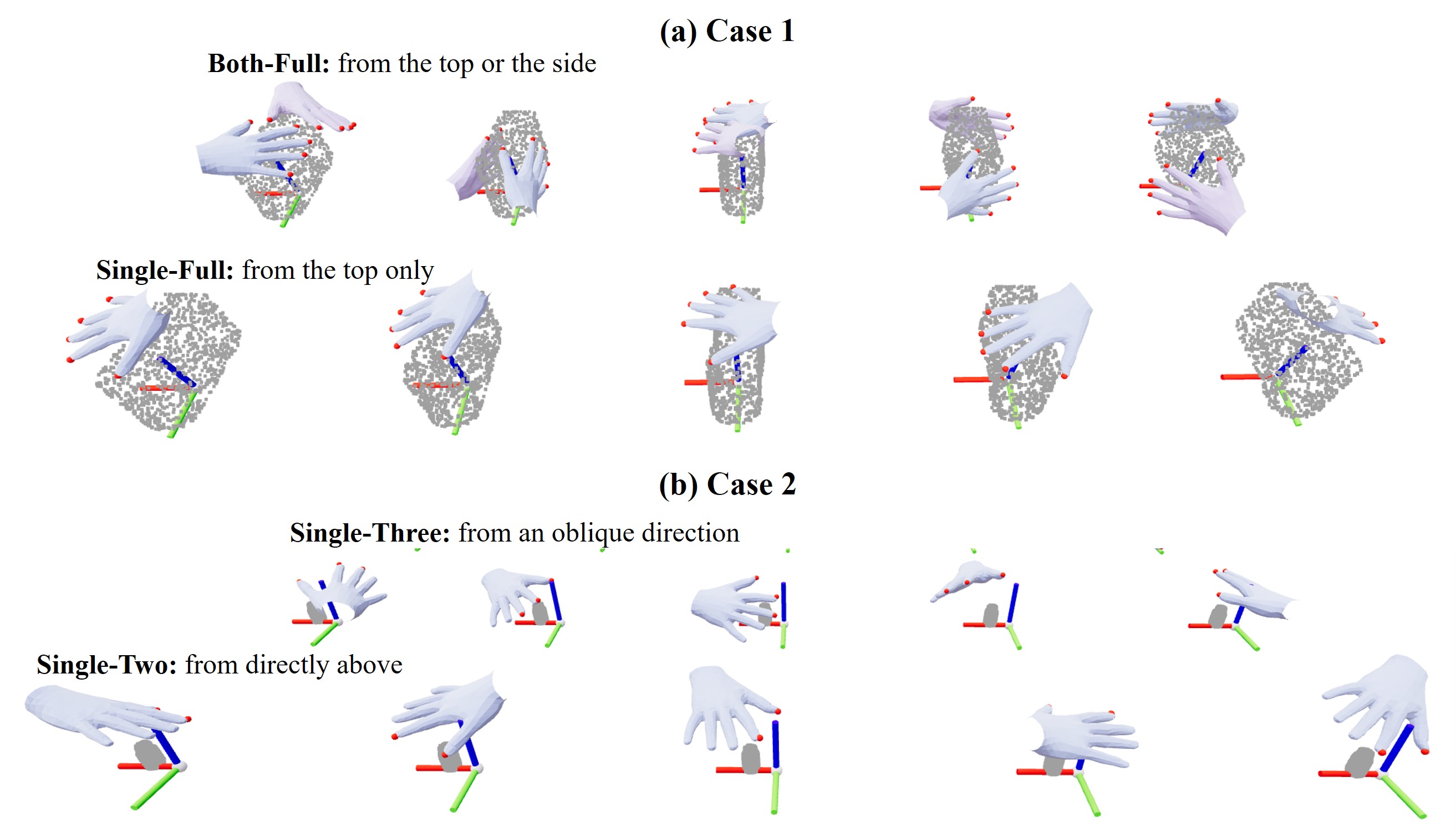}
    \caption{\textbf{Object-conditioned wrist-pose samples for different contact modes.} The examples show that the learned pose prior predicts wrist-pose distributions conditioned jointly on object geometry and contact mode. For a given object geometry, different contact modes favor different approach regions: in Case 1, \textit{Both-Full} admits both top and side bimanual approaches, while \textit{Single-Full} mainly concentrates on top-down approaches; in Case 2, \textit{Single-Three} favors oblique approaches, whereas \textit{Single-Two} favors more direct top-down approaches.}
    \label{fig:human_prior_contact_mode_with_wrist_pose}
\end{figure}

\textbf{Contact-Mode Prior Visualization.}
Fig.~\ref{fig:human_prior_contact_mode} visualizes contact-mode distributions predicted by the learned human prior for diverse object point clouds.
The examples show that the prior adapts its mode probabilities according to object geometry and scale, assigning high probability to sparse single-hand modes for small objects and to full-hand or bimanual modes for larger objects.

\textbf{Wrist-Pose Prior Visualization.}
Fig.~\ref{fig:human_prior_contact_mode_with_wrist_pose} shows mode-conditioned wrist-pose samples from the human prior.
For the same object, different contact modes can induce different preferred approach regions.
In Case 1, the object is large and wide.
Bimanual grasps can approach either from the side or from above because the two hands can jointly cover the object, whereas single-hand side grasps would place the contact region far from the object's center of mass. Therefore, the single-hand prior favors top-down grasps whose contact region is closer to the gravity line through the center of mass.
In Case 2, the object is approximately cylindrical with height larger than its radius.
From a top-down approach, the small radius leaves limited space around the object, making two-finger grasps more suitable; from an oblique side approach, the object height provides enough accessible surface for three-finger contacts, so the \textit{Single-Three} prior favors oblique wrist poses.

\begin{figure}[t]
    \centering
    \includegraphics[width=0.7\linewidth]{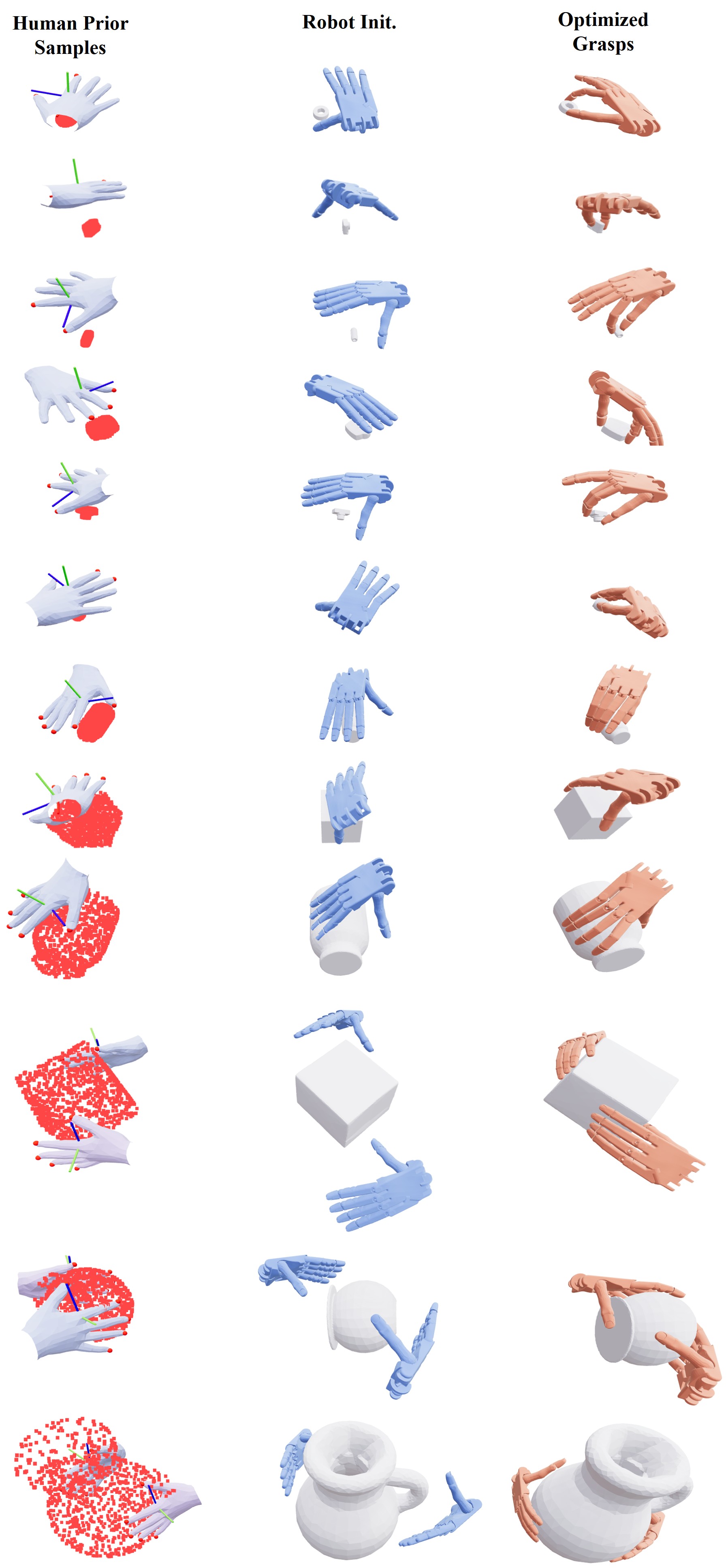}
    \caption{\textbf{From human prior samples to optimized robot grasps.} Left: contact-mode and wrist-pose samples predicted by the human prior for each object. Middle: robot wrist initializations obtained by transferring the prior samples to the robot hand. Right: final robot grasps after force-closure-aware optimization.}
    \label{fig:prior_to_init_to_grasp}
\end{figure}

\textbf{From Human Prior to Robot Grasps.}
Fig.~\ref{fig:prior_to_init_to_grasp} visualizes how the learned human prior guides robot grasp synthesis.
The sampled human-prior wrist poses provide coarse object-conditioned guidance, which is transferred to robot wrist initializations and then refined by force-closure-aware optimization.
The examples cover both single-hand and bimanual grasps across different object scales.

\begin{figure}[!htbp]
    \centering
    \includegraphics[width=\linewidth]{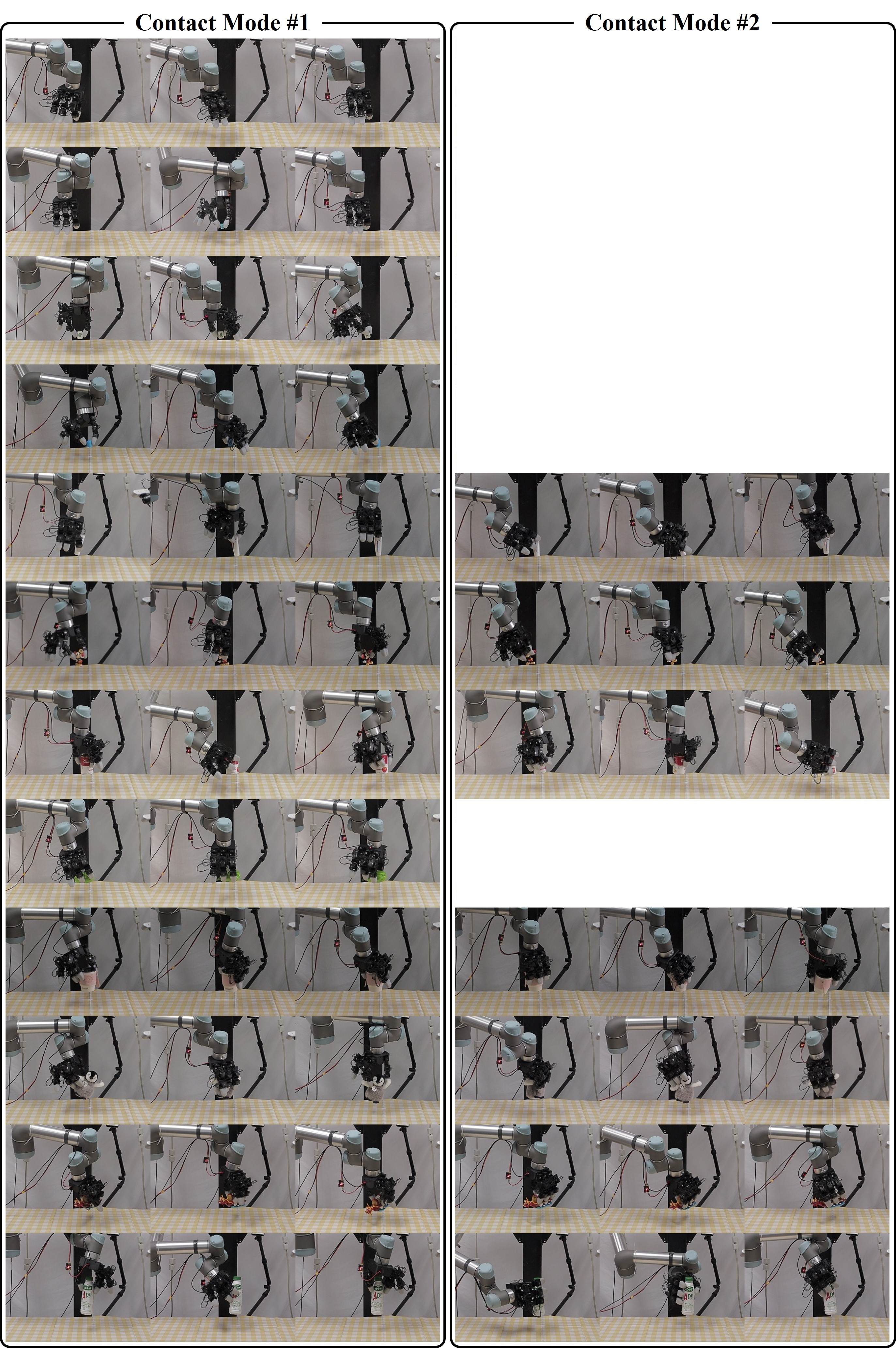}
    \caption{\textbf{Snapshots of real-world grasp on objects \#1--\#12.} Objects are approximately ordered by increasing scale from top to bottom. Some objects allow only one feasible contact mode, whereas others allow two. For each feasible contact mode, we execute three distinct grasps.}
    \label{fig:real_grasp_1}
\end{figure}

\begin{figure}[!htbp]
    \centering
    \includegraphics[width=\linewidth]{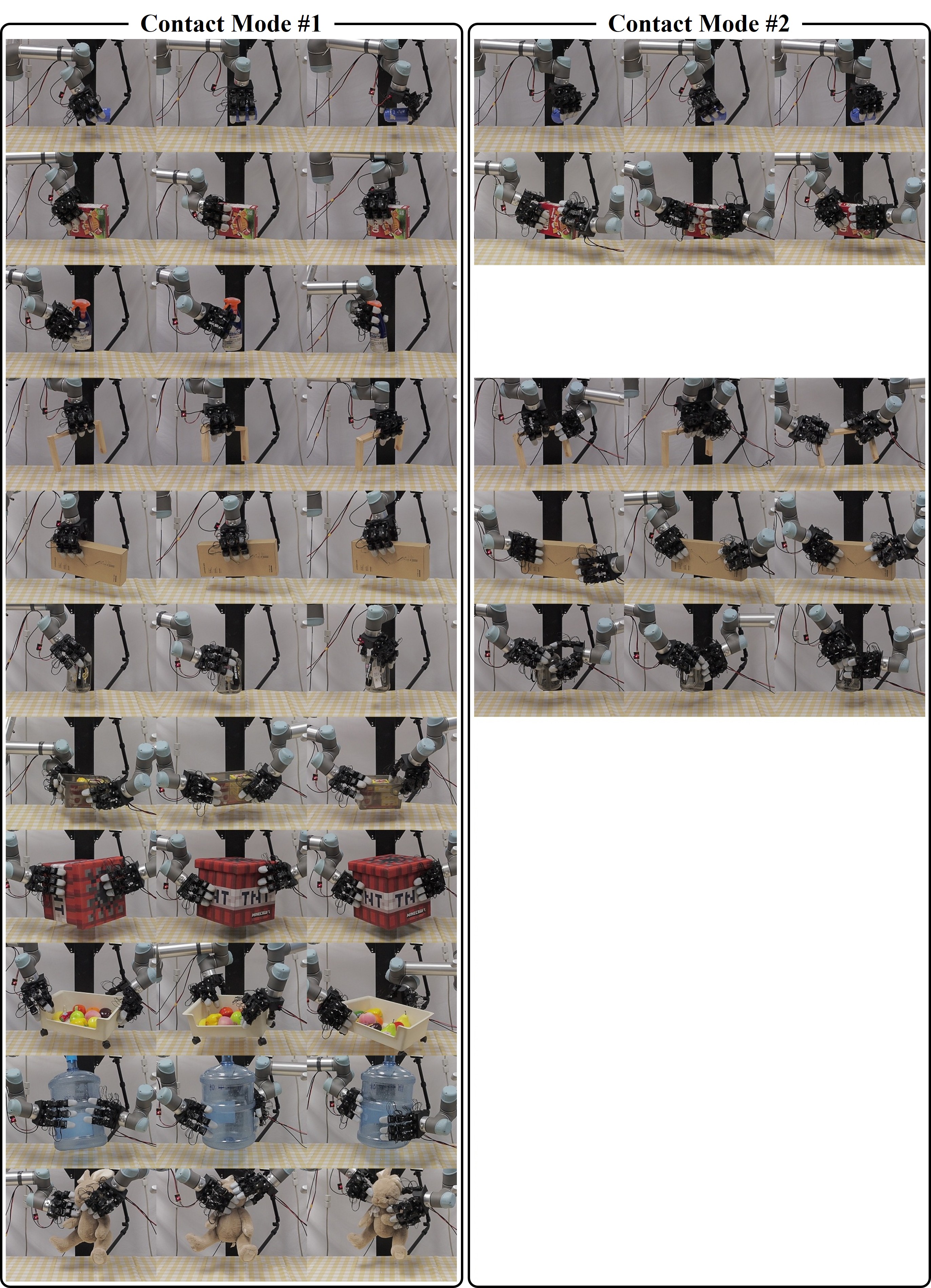}
    \caption{\textbf{Snapshots of real-world grasp on objects \#13--\#23.}}
    \label{fig:real_grasp_2}
\end{figure}

\textbf{Real-World Grasping Snapshots.}
We provide real-world grasping snapshots in Fig.~\ref{fig:real_grasp_1} and Fig.~\ref{fig:real_grasp_2}.
The objects cover a wide range of scales and are approximately ordered from smaller to larger instances.
Depending on the object geometry, each object may support one or two contact modes.
For each feasible contact mode, we execute three distinct grasps to illustrate the diversity of generated bimanual grasp behaviors.

\clearpage
\newpage

\section{Failure Analysis}
\label{app:failure_cases}

\subsection{Failure Analysis of Human-Prior-Guided Grasp Synthesis}
\label{app:synthesis_failure_analysis}

\begin{figure}[!h]
    \centering
\includegraphics[width=0.8\linewidth]{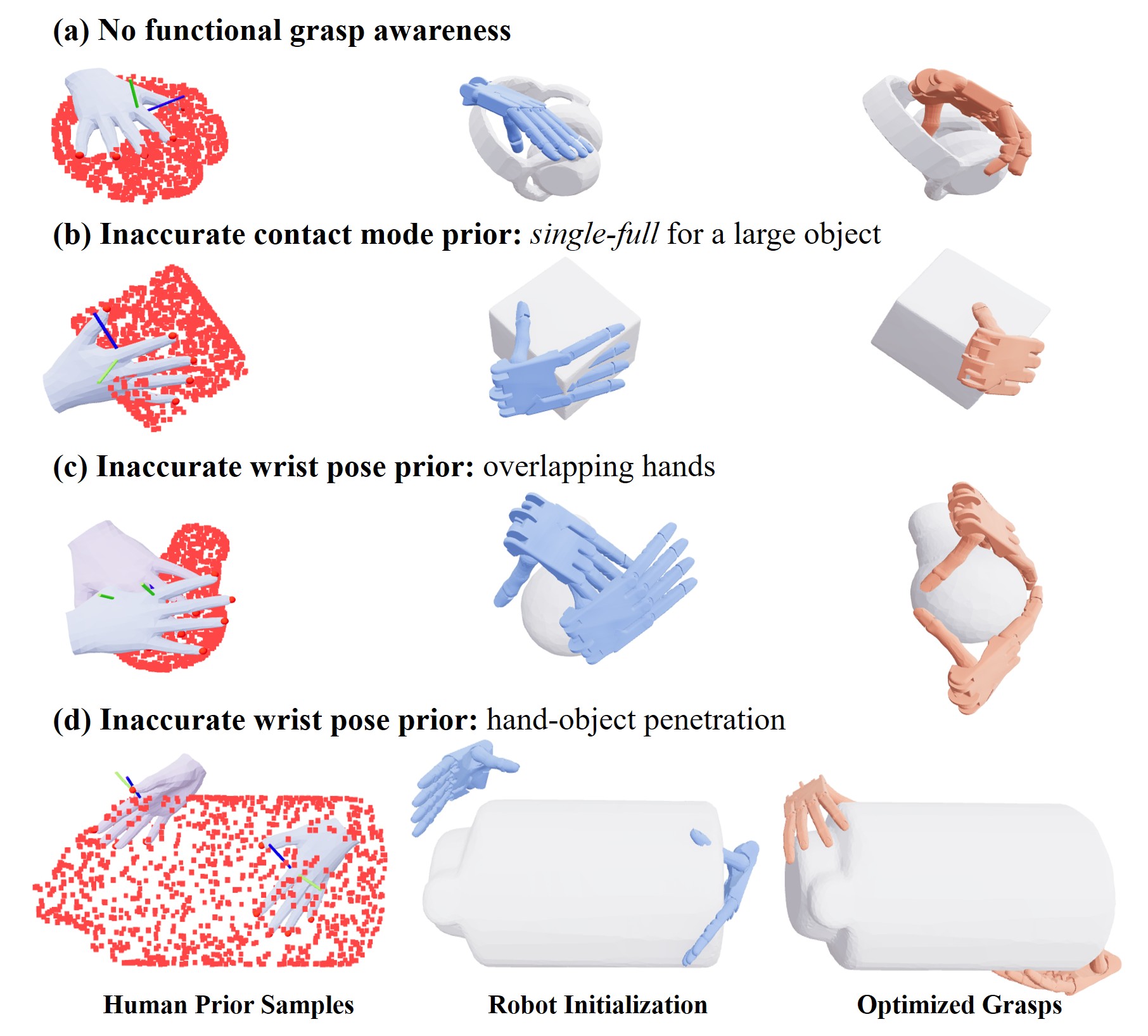}
    \caption{\textbf{Failure analysis of human-prior-guided grasp synthesis.} Each row shows a failure mode from human prior samples to robot initialization and optimized grasps. (a) The human prior and the optimization objective lack functional grasp awareness, so the pipeline may produce geometrically plausible but functionally weak grasps. (b) The contact-mode prior can sometimes predict \textit{Single-Full} for objects that are slightly beyond the capability of single-hand grasping. (c) For relatively small objects, the \textit{Both-Full} prior can sometimes produce partially overlapping hands. (d) For extremely large objects, the \textit{Both-Full} prior can sometimes produce clear hand-object penetration. The hand-hand overlap and hand-object penetration cases can often be mitigated during optimization.}
    \label{fig:synthesis_failure_analysis}
\end{figure}

Fig.~\ref{fig:synthesis_failure_analysis} summarizes typical failure modes observed during human-prior-guided grasp synthesis.
Although the learned prior provides useful global guidance, it remains a coarse prior.
First, the human prior and the optimization objective do not explicitly model functional grasp awareness, so the pipeline may produce grasps that are geometrically plausible but functionally weak.
Second, the contact-mode prior can sometimes predict \textit{Single-Full} for objects that are slightly beyond the capability of single-hand grasping.
Third, for relatively small objects, the \textit{Both-Full} prior can produce partially overlapping hands.
Finally, for extremely large objects, the \textit{Both-Full} prior can produce noticeable hand-object penetration.
As described in Sec.~\ref{app:grasp_synthesis_details}, the bimanual initialization offset helps reduce such problematic initial states.
These hand-hand overlap and hand-object penetration cases can be further mitigated by the feasibility-constrained physical optimization stage. This highlights the benefit of our human-prior-plus-robot-optimization design: the learned human prior provides diverse and global grasp hypotheses, while the robot optimization stage explicitly enforces physical plausibility and adapts the grasps to the target robot embodiment.
These failure modes reveal limitations of the current prior and highlight the need for larger and more comprehensive human grasp datasets, together with stronger functional reasoning for grasp synthesis.

\clearpage

\subsection{Failure Analysis of Distilled Robot Grasp Generators.}

\begin{figure}[t]
    \centering
    \includegraphics[width=\linewidth]{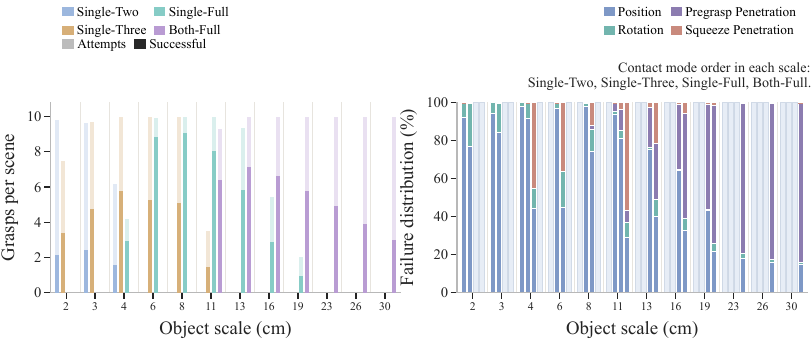}
    \caption{\textbf{Failure distribution of the distilled grasp generator on the Shadow Hand.} Left: number of generated and successful grasps per scene across object scales and contact modes. Right: distribution of failure causes for each contact mode at each object scale, ordered as \textit{Single-Two}, \textit{Single-Three}, \textit{Single-Full}, and \textit{Both-Full}. Failures are categorized as position error (unsuccessful lifting), rotation error (excessive object rotation), pregrasp penetration, and squeeze-stage self-penetration.}
    \label{fig:distill_failure_distribution}
\end{figure}

Fig.~\ref{fig:distill_failure_distribution} further analyzes generator failures after distillation by reporting both the simulation success rates of generated grasps and the distribution of failure causes across object scales and contact modes.
The results reveal several scale- and mode-dependent failure patterns.
First, most failures are caused by position error, where the generated grasp does not successfully lift the object.
This indicates a remaining limitation in the accuracy of grasp-pose generation.
For Single-Full grasps, a noticeable portion of failures comes from squeeze-stage self-penetration, which is not explicitly considered in the current squeeze-pose synthesis process \cite{chen2025bodex}. This issue is particularly pronounced on the Shadow Hand, whose relatively small inter-finger spacing makes self-penetration more likely during finger closure. We plan to address this limitation in future work.
For large objects, especially under the Both-Full mode, many failures already occur at the pregrasp stage due to hand-object penetration, suggesting that the generated global poses are not sufficiently accurate for large-scale objects.
These observations indicate that, although HUGS-synthesized data can effectively supervise online generators, designing robust robot grasp generators across scales and contact modes remains an important direction for future research.

\subsection{Failure Analysis of Real-World Grasping}
\label{app:real_world_failure_cases}

\begin{figure}[htbp]
    \centering
    \includegraphics[width=0.9\linewidth]{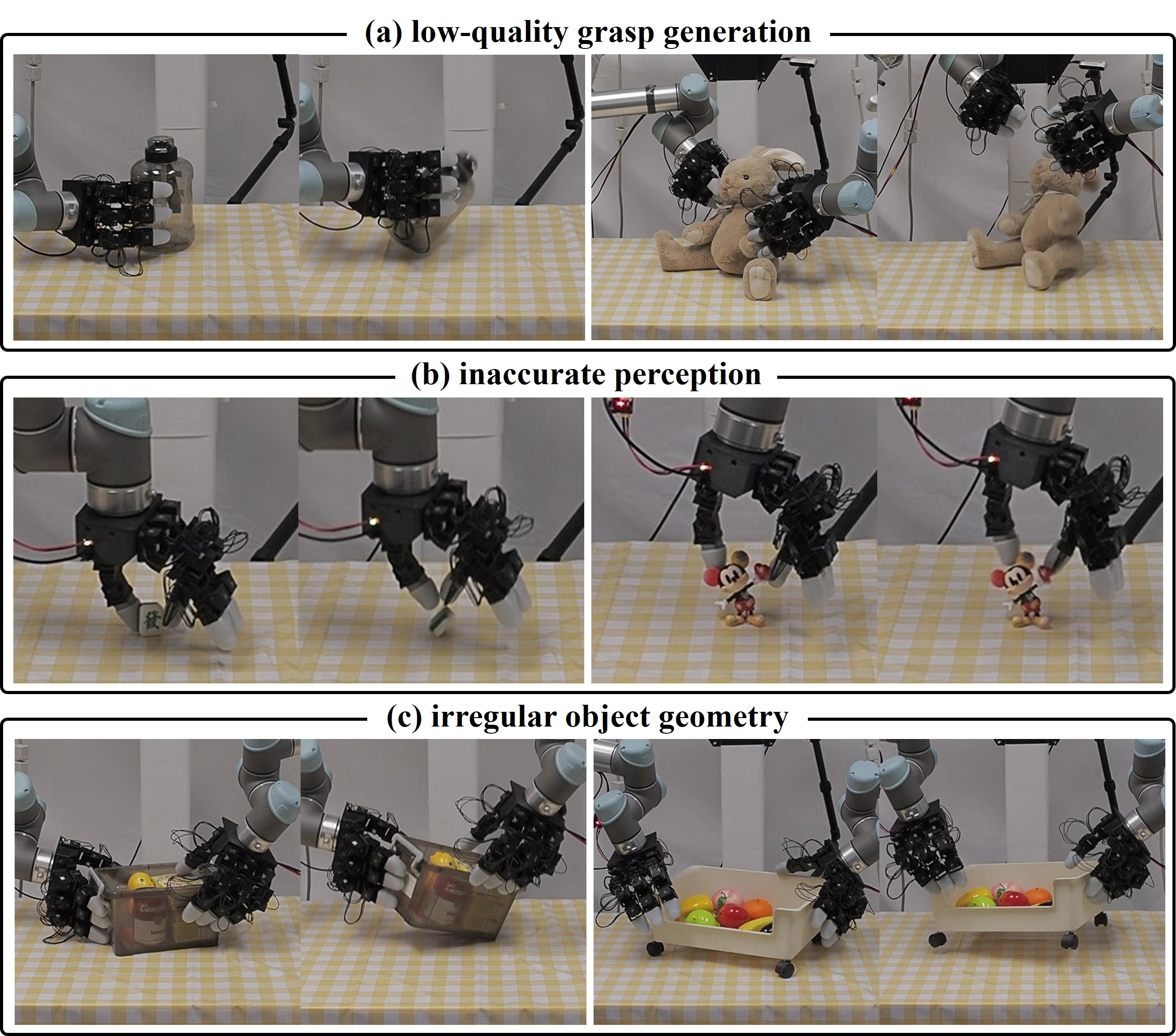}
    \caption{\textbf{Typical failure modes of real-world grasping.} (a) Low-quality grasp generation may produce loose or unstable grasps, suggesting that the current grasp generation network can be further improved. (b) Inaccurate calibration, segmentation, and depth sensing reduce the reliability of reconstructed point clouds, resulting in grasps that miss the object. This issue is particularly pronounced for small objects. (c) Irregularly shaped objects often suffer from poor point-cloud reconstruction and may lie outside the distribution of the training object set, leading to degraded grasp generation. This suggests that the diversity and coverage of the synthetic object dataset can be further improved to enhance generalization.}
    \label{fig:real_world_failure_cases}
\end{figure}

In addition to the generative-network limitations analyzed in the simulation evaluation, real-world failures are affected by several deployment-specific factors, as illustrated in Fig.~\ref{fig:real_world_failure_cases}.
First, low-quality grasp predictions can produce loose grasps that fail to securely lift the object.
Second, calibration, segmentation, and depth reconstruction errors can corrupt the reconstructed object point cloud, causing the generated grasp to miss the target object, especially for small objects.
Third, geometrically irregular objects and challenging cross-scale cases may lie outside the training distribution, leading to degraded grasp generation.
These observations highlight the need for advances at both the model and system levels. On the model side, future work should improve the accuracy and generalization of grasp generators to more diverse object distributions. On the system side, robust real-world deployment will require end-to-end closed-loop execution, tactile adaptation, and collision-aware motion planning.

